\documentclass{article}
\PassOptionsToPackage{numbers}{natbib}
    \usepackage[preprint]{neurips_2020}
\usepackage[utf8]{inputenc} 
\usepackage[T1]{fontenc}    
\usepackage{hyperref}       
\usepackage{url}            
\usepackage{booktabs}       
\usepackage{amsfonts}       
\usepackage{nicefrac}       
\usepackage{microtype}      
\usepackage{graphicx}
\usepackage{multirow}
\usepackage{xcolor}
\usepackage{xspace}
\usepackage{epsfig}
\usepackage{amsmath}
\usepackage{amssymb}
\usepackage{subcaption}
\usepackage{etoolbox}
\usepackage{floatrow}
\usepackage{sidecap}
\usepackage{subfiles}
\newfloatcommand{capbtabbox}{table}[][\FBwidth]

\newcommand{\MyMapTemplatePrefixc}[4]{\expandafter#1\csname#3#4\endcsname{#2{#4}}} 
\forcsvlist{\MyMapTemplatePrefixc {\def} {\mathcal}{c}} {A,B,C,D,E,F,G,H,I,J,K,L,M,N,O,P,Q,R,S,T,U,V,W,X,Y,Z} 

\newcommand{\MyMapTemplatePrefixtb}[5]{\expandafter#1\csname#4#5\endcsname{#2{#3{#5}}}}
\forcsvlist{\MyMapTemplatePrefixtb {\def} {\tilde}{\mathbf}{t}} {A,B,C,D,E,F,G,H,I,J,K,L,M,N,O,P,Q,R,S,T,U,V,W,X,Y,Z}  
\forcsvlist{\MyMapTemplatePrefixtb {\def} {\tilde}{\mathbf}{t}} {0,1,a,b,c,d,e,f,g,h,i,j,k,l,m,n,o,p,q,r,s,u,v,w,x,y,z} 

\newcommand{\MyMapTemplateNoPrefix}[3]{\expandafter#1\csname#3\endcsname{#2{#3}}}
\forcsvlist{\MyMapTemplateNoPrefix {\def} {\mathbf} } {0,1,a,b,c,d,e, f, g, h, i, j, k, l, m, n, o, p, q, r, u, v, w, x, y, z} 
\forcsvlist{\MyMapTemplateNoPrefix {\def} {\mathbf} } {A,B,C,D,E,F,G,H,I,J,K,L,M,N,O,P,Q,R,S,T,U,V,W,X,Y,Z}

\def\E{\mathbb{E}}

\title{Analogical Image Translation for Fog Generation}
\author{
  Rui Gong\textsuperscript{1}, Dengxin Dai\textsuperscript{1}, Yuhua Chen\textsuperscript{1}, Wen Li\textsuperscript{3}, Luc Van Gool\textsuperscript{1,2}\\
  \textsuperscript{1}Computer Vision Lab, ETH Zurich.\\
  \textsuperscript{2}VISICS, KU Leuven.\\
  \textsuperscript{3}University of Electronic Science and Technology of China.\\
  \texttt{\{gongr, dai, yuhua.chen, vangool\}@vision.ee.ethz.ch}\\
  \texttt{liwenbnu@gmail.com}\\
}

\begin{document}

\maketitle

\begin{abstract}
    Image-to-image translation is to map images from a given \emph{style} to another given \emph{style}. While exceptionally successful, current methods assume the availability of training images in both source and target domains, which does not always hold in practice. Inspired by humans' reasoning capability of analogy, we propose analogical image translation (AIT). Given images of two styles in the source domain: $\mathcal{A}$ and $\mathcal{A}^\prime$, along with images $\mathcal{B}$ of the first style in the target domain, learn a model to translate $\mathcal{B}$ to $\mathcal{B}^\prime$ in the target domain, such that $\mathcal{A}:\mathcal{A}^\prime ::\mathcal{B}:\mathcal{B}^\prime$.  AIT is especially useful for translation scenarios in which training data of one style is hard to obtain but training data of the same two styles in another domain is available. For instance, in the case from normal conditions to extreme, rare conditions, obtaining real training images for the latter case is challenging but obtaining synthetic data for both cases is relatively easy. In this work, we are interested in adding adverse weather effects, more specifically fog effects, to images taken in clear weather. To circumvent the challenge of collecting real foggy images, AIT learns with synthetic clear-weather images, synthetic foggy images and real clear-weather images to add fog effects onto real clear-weather images without seeing any real foggy images during training. AIT achieves this zero-shot image translation capability by coupling a supervised training scheme in the synthetic domain, a cycle consistency strategy in the real domain, an adversarial training scheme between the two domains, and a novel network design. Experiments show the effectiveness of our method for zero-short image translation and its benefit for downstream tasks such as semantic foggy scene understanding.     
\end{abstract}

\section{Introduction}
Image-to-image translation has enjoyed tremendous progress in the last years. Excellent methods have been developed for a diverse set of learning paradigms such as supervised learning \cite{isola2017image}, unsupervised learning \cite{zhu2017unpaired,huang2018munit} and few-shot learning \cite{Liu_2019_ICCV}.  While exceptionally successful, current methods have a shared assumption that training data, be it paired or unpaired, is available for both \emph{styles}~\footnote{We reserve `domains' for analogy and use `styles' instead for image translation.}. This may limit the use of image translation when data in one of the two \emph{styles} is hard to obtain, e.g. translation from a normal condition to an extreme, corner-case condition. To address this, we take a new route and propose analogical image translation (AIT) which learns image translation via analogy. 

Analogy is a basic reasoning process to transfer information or meaning from the source to the target. Humans use it commonly to solve problems, provide explanations and make predictions~\cite{image:analogy}. In this paper, we explore the use of analogy as a means for extracting the \emph{gist} of image translation in the source domain and apply it the target domain. Particularly, we aim to solve the following problem: 

\textbf{Problem} (``Analogical Image Translation''): Given images of two styles in the source domain: $\mathcal{A}$ and $\mathcal{A}^\prime$, along with images $\mathcal{B}$ of the first style in the target domain, learn the \emph{translation gist} and apply it to $\mathcal{B}$ to obtain $\mathcal{B}^\prime$, such that $\mathcal{A}:\mathcal{A}^\prime ::\mathcal{B}:\mathcal{B}^\prime$.   

\begin{figure}[t]
\centering
\includegraphics[width=\linewidth]{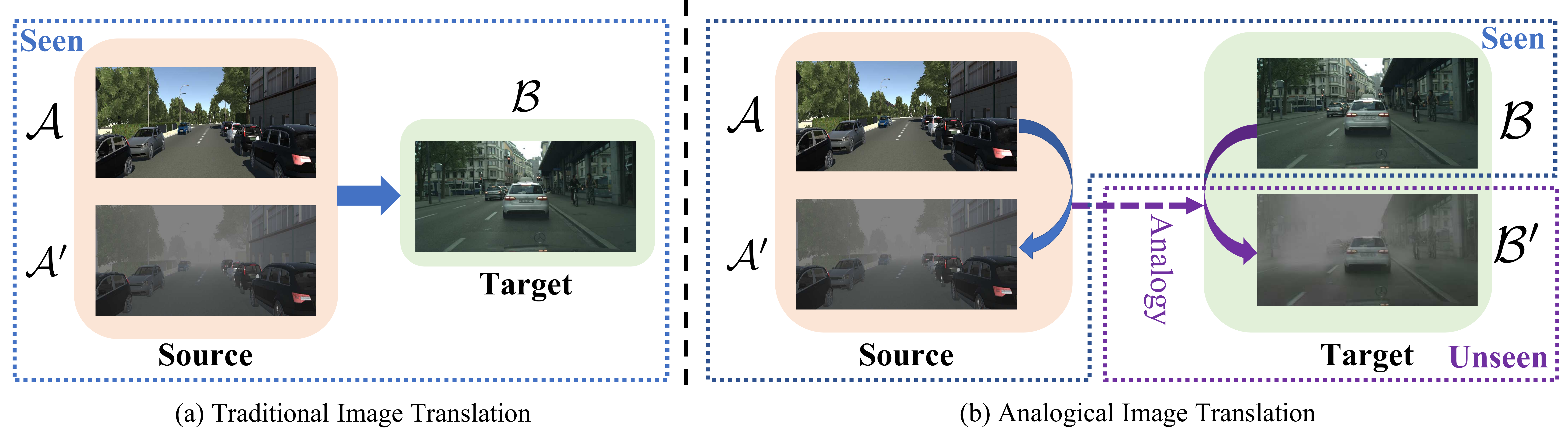}
\caption{\textbf{Traditional image translation v.s. analogical image translation.} Given  images of two styles in the source domain $\cA$ and $\cA^\prime$ along with images of the first style $\cB$ on the target domain, traditional translation methods can only translate between the seen styles $\cA, \cA^\prime$ and $\cB$. The proposed analogical image translation is able to translate $\cB$ to $\cB^\prime$, such that $\mathcal{A}:\mathcal{A}^\prime ::\mathcal{B}:\mathcal{B}^\prime$, without seeing any samples from $\cB^\prime$ at the training and testing stages.}
\label{intervsinner}
\end{figure}

A schematic comparison of AIT to the standard image translation can be found in Fig.~\ref{intervsinner}. 
Our work is partially motivated by the difficulty in obtaining real training images for semantic understanding tasks of autonomous driving in adverse conditions, \emph{e.g.}, the foggy weather.
Despite tremendous progress being made, prior works in semantic scene understanding~\cite{ronneberger2015u, chen2017deeplab, yu2015multi, zhao2017pyramid, lin2017refinenet} have mostly focused on the clear-weather condition, leading to degraded performance for adverse conditions~\cite{halder2019physics, sakaridis2018semantic, blum2019fishyscapes, li2017photo}. Collecting large-scale training datasets for these adverse  conditions and other corner cases may resolve the issue but is hardly scalable and affordable.  

To address this, recent works focus on synthesizing fog effects onto existing clear-weather images by using a physical optical model~\cite{sakaridis2018semantic, hahner2019semantic, ren2016single}. The success of these methods hinges on accurate depth estimation and accurate atmospheric light estimation, both of which, however, are still open problems on their own. Therefore, the synthesized fog still suffers from the presence of artifacts. On the other hand, synthetic foggy images can be generated easily in virtual environments these days~\cite{Gaidon:Virtual:CVPR2016}. This situation motivates the development of our AIT method which learns with the abundant synthetic clear-weather and synthetic foggy images to perform an analogical image translation from real clear-weather images to `real' foggy images. It learns the correlation between synthetic clear-weather and synthetic foggy images, and then applies such learned knowledge to the real domain. We call this learned correlation the \emph{gist} of translation and assume it transferable across domains. Our method is built on top of CycleGAN, so it is named AnalogicalGAN.

AnalogicalGAN achieves this zero-shot translation ability by coupling a supervised training scheme in the synthetic domain, a cycle consistency strategy in the real domain, an adversarial training scheme between the two domains, and a novel network design. More specifically, in the synthetic domain, the \emph{gist} of translation is learned in the supervised manner with the accessible paired
clear-weather and foggy images. Then, this translation \emph{gist} is transferred to the real domain through an adversarial learning scheme. In the real domain, the learning is further supervised through a cycle consistency scheme. The pipeline of AnalogicalGAN can be found in Fig.~\ref{fig:network}. While some choices in AnalogicalGAN are made specifically for fog generation, the method itself has the potential to be used for other AIT tasks.   

Experiments show that AnalogicalGAN outperforms standard image translation methods for our zero-shot image translation task. The superior quality of generated fog is also validated by the state-of-the-art performance of a downstream task semantic foggy scene understanding.   

The rest of the paper is structured as follows: we provide an overview of related work in Sec.~\ref{sec:related}; then we describe our method in Sec.~\ref{sec:method}, which is followed by the experiments in Sec.~\ref{sec:experiemnt}; Finally, we conclude with Sec.~\ref{sec:conclusion}.

\section{Related Works}
\label{sec:related}
\textbf{Image-to-Image Translation}. Image translation methods have been developed to convert images of one given style to another given style with remarkable success in the last years~\cite{zhu2017unpaired,huang2018munit,Liu_2019_ICCV}. Image translation is also becoming a standard step for domain adaptation methods~\cite{Tsai_adaptseg_2018, Lian_2019_ICCV, tsai2019domain, Zou_2019_ICCV, vu2018advent, hoffman2016fcns, chen2018road} -- synthetic images are first translated to `real' images on which the downstream tasks such as segmentation and detection are then conducted~\cite{Hoffman_cycada2017, chen2019learning, li2019bidirectional, gong2019dlow, dundar2018domain}. The standard image translation framework \cite{zhu2017unpaired, isola2017image, huang2018munit, liu2017unsupervised} requires the availability of images of both styles involved in the translation. This limits the use of this framework. Our proposed AIT method enriches the learning paradigms by exploring the use of analogy, and opens a new avenue for leveraging synthetic data for real-world applications. The closest work to ours is PuppetGAN~\cite{puppetgan}, which learns to manipulate individual visual attributes of objects in a real scene using examples of attribute manipulation in a simulation. While the high-level spirit is similar, our work differs significantly from PuppetGAN. PuppetGAN focuses on manipulating attributes like \emph{pose} and \emph{size} of faces and digits while we focus on adding weather effects to general outdoor images. This difference also leads to different algorithms.      

\textbf{Semantic Foggy Scene Understanding}.
Our work is also related to methods of semantic foggy scene understanding (SFSU). The SFSU task aims to improve the  performance of semantic scene understanding under foggy  condition~\cite{sakaridis2018semantic, dai2019curriculum, hahner2019semantic, erkent2020semantic, tarel2010improved}. Due to the difficulty of gathering and labeling large-scale foggy weather image dataset, some works \cite{sakaridis2018semantic, dai2019curriculum} propose to synthesize fog by applying the physical model to the real clear weather images in the Cityscapes dataset \cite{cordts2016cityscapes}, leading to the Foggy Cityscapes dataset. While yielding improved results, these methods need accurate depth estimation and atmospheric light estimation. Failures of the two tasks will result in notorious artifacts. Unlike these works, our proposed AIT method does not rely on estimated depth and atmospheric light from real-world images, and is able to make full use of the abundant synthetic data.

\textbf{Unsupervised Domain Adaptation}. Our work also shares similarity with unsupervised domain adaptation (UDA). UDA has been extensively studied in the past years, mainly for semantic segmentation~\cite{chen2018road,tsai2019domain,dai2019curriculum,li2019bidirectional} and object detection~\cite{chen2018domain,xie2019multi,zhu2019adapting}. Given a set of images and the corresponding annotations in the source domain, along with a set of unlabeled images in the target domain, the goal is to learn a semantic model which can perform well in the target domain as well. Our AIT shares the same spirit by transferring the learned function from the source domain to the target domain without using annotations (images of desired styles) in the target domain. Our work tackles a different task than those UDA methods did, which leads to a different algorithm and different applications.   

\section{Analogical Image Translation}
\label{sec:method}
\subsection{Problem Statement
}\label{sec:pb_st}
In the image translation problem, we are given a source domain $\cS$ and a target domain $\cT$, which consist of the samples $\x^s \in \cS$ and $\x^t \in \cT$, respectively. The goal of traditional image translation is to transfer image samples $\x^s$ and $\x^t$ between domain $\cS$ and domain $\cT$. In our work, we propose analogical image translation (AIT), where the source domain $\cS$ and the target domain $\cT$ cover two styles $\cA, \cA^\prime$ and $\cB, \cB^\prime$, respectively. But during training and testing, there are only samples $\x^{a}\in \cA$, $\x^{a^\prime}\in \cA^\prime$ and $\x^{b}\in \cB$ available. AIT aims to translate from available samples $\x^{a}, \x^{a\prime}, \x^{b}$ to the unseen samples $\x^{b^\prime}$, such that $\x^{a}:\x^{a^\prime}::\x^{b}:\x^{b^\prime}$. The data distribution is denoted as $\x^{a}\sim P_{A}, \x^{a^\prime}\sim P_{A^\prime}, \x^{b}\sim P_{B}$ and $\x^{b^\prime}\sim P_{B^\prime}$. 

While the previous works \cite{zhu2017unpaired, Hoffman_cycada2017, huang2018munit, dundar2018domain} focus on learning the mapping $G_{ST}: \cS\rightarrow\cT$, our objective in this work is to learn the mapping $G_{BB^\prime}: \cB\rightarrow \cB^\prime$ conditioned on the mapping $G_{AA^\prime}: \cA\rightarrow \cA^\prime$. 

\subsection{AnalogicalGAN Model}
\begin{SCfigure}
\includegraphics[width=0.6\linewidth]{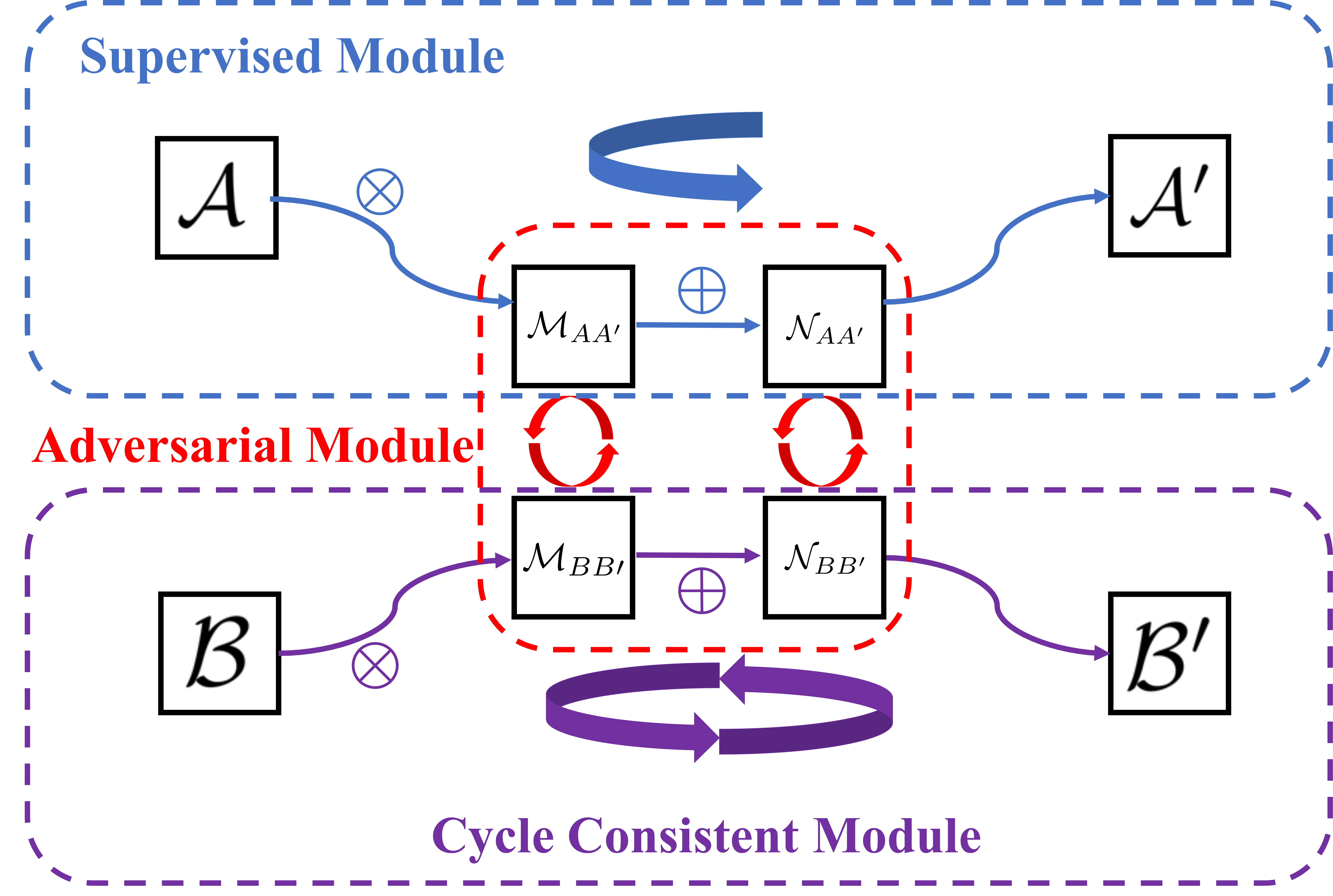}
\vspace{-10pt}
\caption{AnalogicalGAN model overview. The AnalogicalGAN model mainly consists of three modules: the supervised module, the adversarial module and the cycle-consistent module. The supervised module is utilized to disentangle the \emph{gist}, $\cM_{AA^\prime}, \cN_{AA^\prime}$, in the supervised way. The adversarial module transfers the \emph{gist} from source domain, $\cM_{AA^\prime}, \cN_{AA^\prime}$, to the target domain, $\cM_{BB^\prime}, \cN_{BB^\prime}$. The cycle consistent module is adopted to make the transferred \emph{gist} to be compatible with the target domain.}
\label{fig:network}
\end{SCfigure}

In this section, we present our AnalogicalGAN model for the analogical image translation problem. The key idea of our AnalogicalGAN model is to disentangle the translation \emph{gist} in the source domain, transfer the \emph{gist} to the target domain, and make the \emph{gist} compatible with the target domain. In our work, the \emph{gist} is measured with the alignment map $\cM$ and the residual map $\cN$, formally denoted as $\{\cM, \cN\}$. Taking the translation direction into account, the $\{\cM, \cN\}$ can be further expressed in detail as $\cM=\{\cM_{AA^\prime}, \cM_{A^\prime A}, \cM_{BB^\prime}, \cM_{B^\prime B}\}, \cN=\{\cN_{AA^\prime}, \cN_{A^\prime A}, \cN_{BB^\prime}, \cN_{B^\prime B}\}$. Moreover, the \emph{gist} is assumed to be invariant to the source domain and the target domain. Then the \emph{gist} can be defined implicitly as: 
\begin{eqnarray}
\label{eq:innerdiff_syn1}
\cA^\prime =&\!\!\!\!\cA\odot \cM_{AA^\prime} + \cN_{AA^\prime}, \\
\label{innerdiff_real1}
\cB^\prime =&\!\!\!\!\cB\odot \cM_{BB\prime} + \cN_{BB^\prime},\\
\label{eq:innerdiff_syn2}
\cA =&\!\!\!\!\cA^\prime\odot \cM_{A^\prime A} + \cN_{A^\prime A}, \\
\label{innerdiff_real2}
\cB =&\!\!\!\!\cB^\prime\odot \cM_{B\prime B} + \cN_{B^\prime B},
\end{eqnarray}
where $\odot$ denotes the element-wise multiplication.
On this basis, as shown in Fig. \ref{fig:network}, taking the direction of first style to second style for example, i.e. $\cA\rightarrow \cA^\prime$, $\cB\rightarrow \cB^\prime$, our framework consists of three main components: the supervised module, the adversarial module and the cycle consistent module. Firstly, on the source domain, due to the paired samples from $\cA$ and $\cA\prime$ available, the \emph{gist}, $\cM_{AA^\prime}, \cN_{AA^\prime}$, is disentangled in the supervised way according to the Eq. (\ref{eq:innerdiff_syn1}), which forms the supervised module. Secondly, in the adversarial module, based on the domain invariant assumption of the \emph{gist}, the \emph{gist} on the source domain, $\cM_{AA^\prime}, \cN_{AA^\prime}$ ,is transferred to the target domain, $\cM_{BB^\prime}, \cN_{BB^\prime}$, through the adversarial learning. Thirdly, on the target domain, due to the unavailability of the second style $\cB^\prime$, the \emph{gist}, $\cM_{BB^\prime}, \cN_{BB^\prime}$ is retained to be compatible with the target domain through the cycle consistency, constructing the cycle consistent module. The other direction from the second style to the first style, $\cA^\prime \rightarrow \cA, \cB^\prime \rightarrow \cB$, acts in the same way. Next, the different modules and corresponding loss function are introduced in detail. 

\textbf{Supervised Module.} The supervised module is used to disentangle the \emph{gist}, $\cM, \cN$, from the source domain. Given the paired sample $\x^{a}\in \cA$ and $\x^{a^\prime}\in \cA^\prime$ on the source domain $\cS$, the translation between $\cA$ and $\cA^\prime$ can be trained in the supervised way, by substituting in Eq.(\ref{eq:innerdiff_syn1}), written as,
\begin{eqnarray}
\cL_{sup} =\E_{\x^{a}\sim{P_{A}}}\left[\|\x^{a}\odot \m^{aa^\prime}+\n^{aa^\prime}-\x^{a^\prime}\|_{1}\right]
+\E_{\x^{a^\prime}\sim{P_{A^\prime}}}\left[\|\x^{a^\prime}\odot \m^{a^\prime a}+\n^{a^\prime a}-\x^a\|_{1}\right],
\end{eqnarray}
where $(\m^{aa^\prime}, \n^{aa^\prime}) = G_{AA^\prime}(\x^{a})$ and $(\m^{a^\prime a}, \n^{a^\prime a}) = G_{A^\prime A}(\x^{a^\prime})$.

\textbf{Adversarial Module.} The adversarial module aims to transfer the \emph{gist}, disentangled from the source domain, to the real domain. Specifically, taking the direction, $\cA\rightarrow \cA^\prime$, $\cB\rightarrow \cB^\prime$, for example, we introduce the discriminator $D_{I}$ to distinguish the \emph{gist} between the source domain, $\{\cM_{AA^\prime}, \cN_{AA^\prime}\}$, and the target domain, $\{\cM_{BB^\prime}, \cN_{BB^\prime}\}$. And the discriminator $D_{J}$ acts in the same way in the inverse direction $\cA^\prime\rightarrow \cA$, $\cB^\prime\rightarrow \cB$. Then the adversarial loss of \emph{gist} $\{\cM, \cN\}$ on $\cS$ and $\cT$ can be written as,
\begin{eqnarray}
\label{eqn:loss_adv}
\cL_{adv}(G_{AA^\prime}, G_{BB^\prime}, D_I) =&\!\!\!\!\!\!\!\!\!\!\!\!\!\!\!\!\!\!\!\E_{\x^a\sim{P_A}}\left[\log(D_{I}(G_{AA^\prime}(\x^a)))\right]\\ \nonumber
+&\!\!\!\!\E_{\x^b\sim{P_B}}\left[\log(1-D_{I}(G_{BB^\prime}(\x^b)))\right].
\end{eqnarray}
The similar adversarial loss $\cL_{adv}(G_{A^\prime A}, G_{B^\prime B}, D_J)$ is also defined for the direction $\cA^\prime\rightarrow \cA$, $\cB^\prime\rightarrow \cB$. Then the \emph{gist} adversarial loss can be formulated as:
\begin{eqnarray}
\cL_{adv}=\cL_{adv}(G_{AA^\prime}, G_{BB^\prime}, D_I) + \cL_{adv}(G_{A^\prime A}, G_{B^\prime B}, D_J).
\end{eqnarray}
In order to make the mapping $G_{BB^\prime}$ conditional on $G_{AA^\prime}$, the $G_{AA^\prime}$ and $G_{BB^\prime}$, $G_{A^\prime A}$ and $G_{B^\prime B}$ share all the parameters, respectively.

\textbf{Cycle Consistent Module.} The cycle consistent module is utilized to make the \emph{gist} compatible with the target domain, i.e., preserve the target domain feature of the translated \emph{gist}. Accordingly, the reconstruction loss is taken to recover $\x^b$ from the translated image $\x^{b^\prime}$ through the inverse mapping $G_{B^\prime B}$. Furthermore, in order to strengthen the recovery, another discriminator $D_{T}$ is introduced to distinguish between the recovered $\x^b$ and the original $\x^b$. Then the image cycle consistency loss $\cL_{cyc}$ consists of the reconstruction loss $\cL_{rec}$ and the adversarial loss $\cL_{adv}(G_{BB^\prime}, G_{B^\prime B}, D_{T})$, by substituing in Eq.~(\ref{innerdiff_real1}), given by:
\begin{eqnarray}
\cL_{cyc} =&\!\!\!\!\!\!\!\!\!\!\!\!\!\!\!\!\!\!\!\!\!\!\!\!\!\!\!\!\!\!\!\!\!\!\!\!\!\!\!\!\!\!\!\!\!\!\!\!\!\!\!\!\!\!\!\!\!\!\!\!\!\!\!\!\!\!\!\!\!\!\cL_{rec} + \cL_{adv}(G_{BB^\prime}, G_{B^\prime B}, D_{T})\\
\cL_{rec} =&\!\!\!\!\!\!\!\!\!\!\!\!\!\!\!\E_{\x^{b}\sim{P_{B}}}\left[\|\m^{b^\prime b}\odot(\m^{bb^\prime}\odot \x^b+\n^{bb^\prime})+\n^{b^\prime b}-\x^{b}\|_{1}\right]\\
\cL_{adv}(G_{BB^\prime}, G_{B^\prime B},D_T) =&\!\!\E_{\x^b\sim{P_B}}\left[\log(1-D_{T}(\m^{b^\prime b}\odot(\m^{bb^\prime}\odot \x^b+\n^{bb^\prime})+\n^{b^\prime b}))\right]\\
 +&\!\!\!\!\!\!\!\!\!\!\!\!\!\!\!\!\!\!\!\!\!\!\!\!\!\!\!\!\!\!\!\!\!\!\!\!\!\!\!\!\!\!\!\!\!\!\!\!\!\!\!\!\!\!\!\!\!\!\!\!\!\!\!\!\!\!\!\!\!\!\!\!\!\!\!\!\!\!\!\!\!\!\!\!\!\!\E_{\x^b\sim{P_B}}\left[\log(D_{T}(\x^b))\right],\nonumber
\end{eqnarray}
where $(\m^{bb^\prime}, \n^{bb^\prime}) = G_{BB^\prime}(\x^{b})$, $(\m^{b^\prime b}, \n^{b^\prime b}) = G_{B^\prime B}(\x^{b^\prime})$ and $\x^{b^\prime} = \m^{bb^\prime}\odot \x^{b}+\n^{bb^\prime}$. 

\textbf{Auxiliary Module.} Besides the three main modules, the auxiliary module is added to assist the analogical image translation process and introduce the auxiliary information. From \cite{huang2018munit} and \cite{johnson2016perceptual}, the perceptual loss calculates the VGG feature distance $\Phi(\cdot)$ \cite{simonyan2014very} between the translated image and the reference image, and is proven to be able to assist the image translation process. Generalizing the perceptual loss to analogical image translation, the perceptual loss is given in the analogical way, formulated as, 
\begin{eqnarray}
\d^S =&\!\!\!\!\!\!\!\!\!\!\!\!\!\!\!\!\!\!\!\!\!\!\!\!\!\!\!\Phi(\x^{a^\prime}) - \Phi(\x^{a}) \\
\d^T =&\!\!\!\!\!\!\!\!\!\!\!\!\!\!\!\!\!\!\!\!\!\!\!\!\!\!\!\Phi(\x^{b^\prime}) - \Phi(\x^{b}) \\
\cL_{percep} =&\!\!\!\E_{\x^{b}\sim{P_{B}}}\left[\|\d^S-\d^T\|_{1}\right],
\end{eqnarray}
where $(\m^{bb^\prime}, \n^{bb^\prime}) = G_{BB^\prime}(\x^b)$ and $\x^{b^\prime} = \m^{bb^\prime}\odot \x^{b}+\n^{bb^\prime}$. 
Meanwhile, in terms of specific setting such as the analogical foggy image translation, the corresponding auxiliary information to fog effects, such as depth information \cite{fattal2008single, sakaridis2018semantic, dai2019curriculum}, can also be leveraged. By introducing the mapping $G_{IH}: \cA\rightarrow \cH_{S}, \cB\rightarrow \cH_{T}$ and $G_{JH}: \cA^\prime\rightarrow \cH_{S}, \cB^\prime\rightarrow \cH_{T}$, where $\cH_{S}$ and $\cH_{T}$ denote the depth domain corresponding to $\cS$ and $\cT$, composed of depth map $\d^{S}$ and $\d^{T}$, respectively. The auxiliary depth loss is given by, 
\begin{eqnarray}
\cL_{dep} = &\E_{\x^{a}\sim{P_{A}}}\left[\|G_{IH}(\x^{a})-\d^S\|_{1}\right] + \E_{\x^{a^\prime}\sim{P_{A^\prime}}}\left[\|G_{JH}(\x^{a^\prime})-\d^S\|_{1}\right]\\
+&\E_{\x^{b}\sim{P_{B}}}\left[\|G_{IH}(\x^{b})-\d^T\|_{1}\right] + \E_{\x^{b^\prime}\sim{P_{B^\prime}}}\left[\|G_{JH}(\x^{b^\prime})-\d^T\|_{1}\right].\nonumber
\end{eqnarray}
By sharing the network parameters between $G_{IH}$, $G_{AA^\prime}$, and $G_{BB^\prime}$, $G_{JH}$, $G_{A^\prime A}$ and $G_{B^\prime B}$ respectively, the depth information is implicitly encoded into our analogical translation process. 

\textbf{Full Objective.} Integrating the losses defined above, our full objective for AnalogicalGAN model can be defined as:
\begin{eqnarray}
\cL = \cL_{adv} + \lambda_{1}\cL_{sup} + \lambda_{2}\cL_{cyc} + \lambda_{3}\cL_{dep} + \lambda_{4}\cL_{percep},
\end{eqnarray}
where $\lambda_{1}, \lambda_{2}$, $\lambda_{3}$ and $\lambda_{4}$ are hyper-parameters used to balance different parts of training loss. Following the general manner for training the adversarial model, the full objective is trained in the minimax way, i.e. minimize the objective for the generator while maximizing the objective for discriminator. 

\textbf{Domain Interpolation.} Benefiting from the disentangled \emph{gist}, our AnalogicalGAN is able to generate the intermediate domain between $\cB$ and $\cB^\prime$ during testing stage. Following \cite{gong2019dlow}, the variable $z\in [0, 1]$ is used to measure the domainness. The intermediate domain between $\cB$ and $\cB^\prime$ are denoted as $\cI_{B}^{(z)}$. When $z=0$, the intermediate domain $\cI_{B}^{(z)}$ are identical to $\cB$; and when $z=1$, it is identical to $\cB^\prime$. In order to generated the intermediate domain, it is assumed that the \emph{gist} between $\cB$ and $\cB^\prime$ is linear. On the basis of the linear assumption and Eq. (\ref{innerdiff_real1}), the intermediate domain can be written as,
\begin{eqnarray}
\cI_{B}^{(z)} = \cB\odot ((\cM_{BB^\prime}-1)\times z + 1) + \cN_{BB^\prime}\times z.
\end{eqnarray}

\section{Experiments}
\label{sec:experiemnt}

In this section, we evaluate our AnalogicalGAN model for fog generation task. As aforementioned, our method consists of two domains: a source domain $\cS$ and a target domain $\cT$. On $\cS$ and $\cT$, there are two styles $\cA$ and $\cA^\prime$, $\cB$ and $\cB^\prime$ defined, respectively. Because training data for $\cB^\prime$ is unavailable, existing image translation methods can only be trained for $\cA^\prime$ and $\cB$, which does not serve the exact purpose -- generating data in $\cB^\prime$. Training standard translation methods on $\cA^\prime$ and $\cB$, nevertheless, can be taken as baseline methods.  
In our experiments, we instantiate $\cS$, $\cT$, $\cA$, $\cA^\prime$, $\cB$ and $\cB^\prime$ as follows: \emph{synthetic} as $\cS$, \emph{real} as $\cT$, \emph{synthetic, clear weather} as $\cA$, \emph{synthetic, foggy weather} as $\cA^\prime$, \emph{real, clear weather} as $\cB$, and \emph{real, foggy weather} as $\cB^\prime$.

\subsection{Analogical Image Translation}\label{sec:inner_trans}
We conduct the analogical image translation experiments by regarding the Virtual KITTI \cite{Gaidon:Virtual:CVPR2016} as synthetic domain, while the Cityscapes \cite{cordts2016cityscapes} as the real domain. The depth maps of the Cityscapes images are generated by the pretrained deep model developed in \cite{chang2018pyramid}. 

\textbf{Virtual KITTI.} Virtual KITTI is a dataset consisting of 2136 photo-realistic synthetic clear weather images imitating the content and structure of KITTI dataset \cite{geiger2013vision}, each of which has paired foggy weather image and corresponding depth map available.

\textbf{Cityscapes.} Cityscapes is a dataset covering 2975 real clear weather images taken from different European cities, which are densely labeled with 19 classes for semantic segmentation.

We follow the training procedure as CycleGAN \cite{zhu2017unpaired}. The learning rate is fixed to 0.0002 and the image is resized to $512\times 256$. The weight of the \emph{gist} adversarial loss is set as 3, the weight of cycle consistency adversarial loss is set as 1, and the weight of rest parts are 10. 

\textbf{Quantitative Results.} In order to validate the effectiveness of our AnalogicalGAN model for the AIT task, a user study on Amazon Mechanical Turk (AMT) is conducted to compare the translation results of our AnalogicalGAN model with the state-of-the-art traditional image translation methods CycleGAN \cite{zhu2017unpaired} and MUNIT \cite{huang2018munit}. In order to guarantee good quality, we only employ AMT Masters in our study. Each individual task completed by the participants, referred to as Human Intelligence Task (HIT), comprises two image pairs to be compared: Ours vs. CycleGAN and ours vs. MUNIT. In total, 100 HITs were used, each is completed by three annotators and the results are averaged. For each image pair, the users were asked to select the image that looks more like a real foggy image.  
In Fig.~\ref{fig:user_study}, the user study results are listed. From the figure, one can see that users prefer our translation results compared to CycleGAN ($61.0\%$ v.s. $39.0\%$) and MUNIT ($66.7\%$ v.s. $33.3\%$). 

\begin{figure}[ht]
\centering
\includegraphics[width=1.0\linewidth]{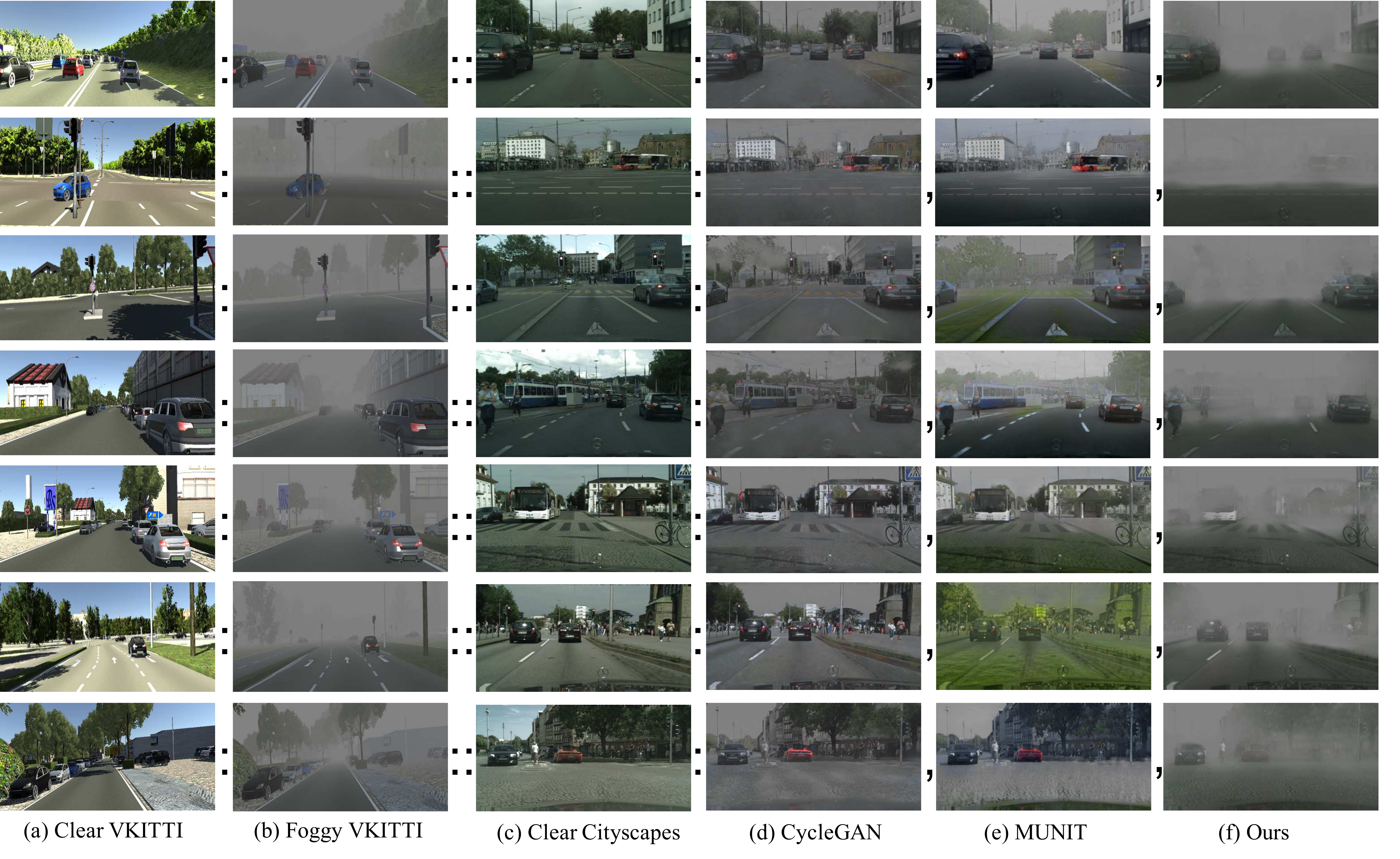}
\caption{Comparison of the analogical translation results of our AnalogicalGAN model (column (f)) with the traditional image translation methods (column (d) and column (e)). The column (a), column (b) and column (c) shows the synthetic clear weather image (Clear Virtual KITTI), the synthetic foggy weather image (Foggy Virtual KITTI) and the real clear weather image (Cityscapes), respectively. The analogical translation is described as, column (a) $:$ column (b) $::$ column (c) $:$ column (d), column (e), column (f).}
\label{fig:quality_comp}
\end{figure}

\textbf{Qualitative Results.} Furthermore, we show the qualitative comparison in Fig. \ref{fig:quality_comp}. From Fig. \ref{fig:quality_comp}, it is observed that the standard image translation models CycleGAN (refer to Fig. \ref{fig:quality_comp}(d)) and MUNIT (refer to Fig. \ref{fig:quality_comp}(e)) suffer from inheriting synthetic features from the Virtual KITTI (refer to Fig. \ref{fig:quality_comp}(b)) such as the color of the car, the lines on the road and the skin of the people. Besides, though the translated foggy part tends to be in gray, it loses the correct sense that fog changes with depth. In contrast, our AnalogicalGAN model, the analogical image translation framework, preserves the real feature of the objects in the scene, generates realistic foggy images and yields the right sense that fog changes with the depth of the scene as shown in Fig. \ref{fig:quality_comp}(f). 

\begin{minipage}{\textwidth}
  \begin{minipage}[b]{0.49\textwidth}
    \centering
    \includegraphics[width=0.8\textwidth]{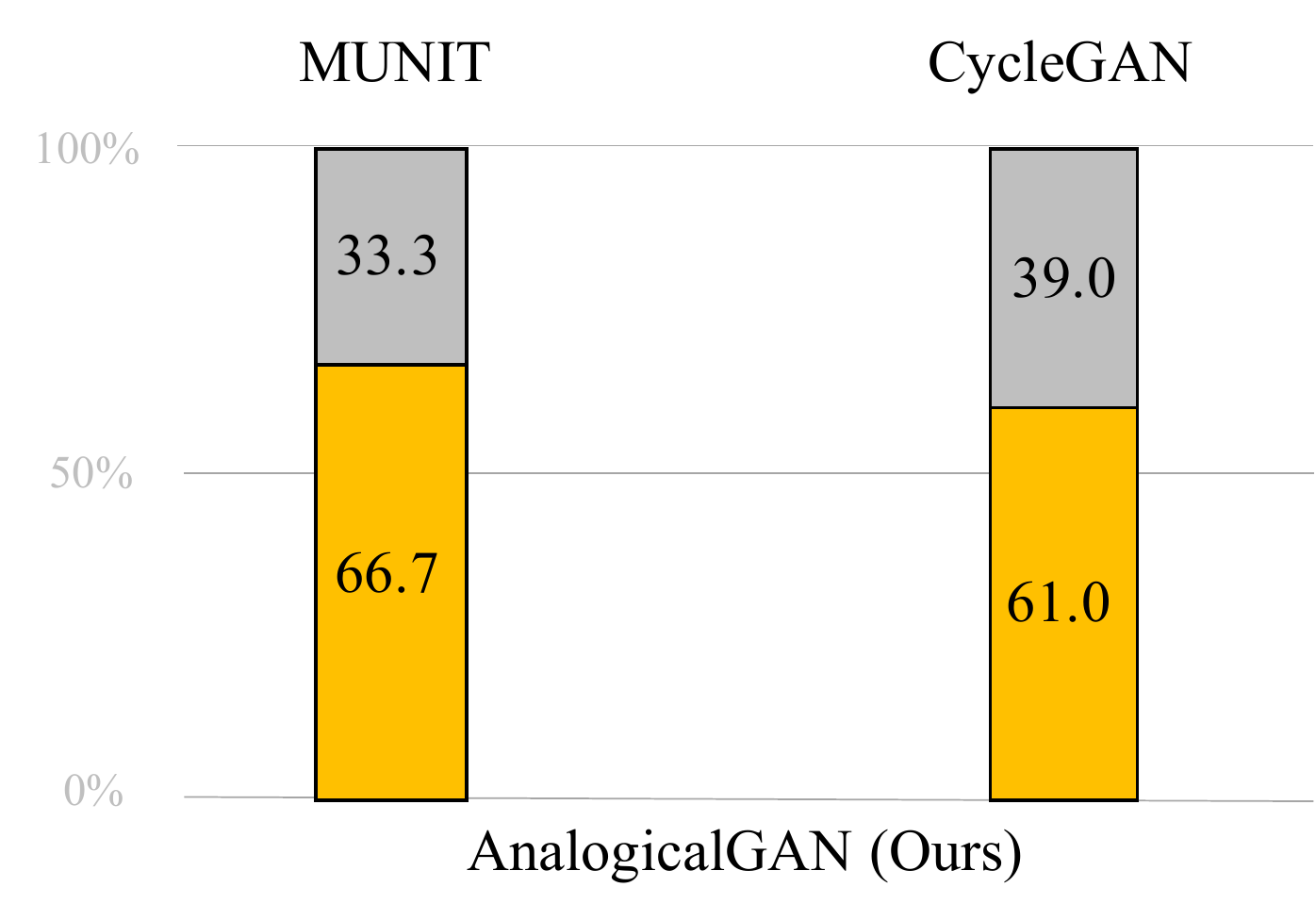}
    \vspace{-10pt}
    \captionof{figure}{User study results for fog generation. It is observed that more users prefer the translation results of our AnalogicalGAN model compared to that of CycleGAN and MUNIT.\label{fig:user_study}}
  \end{minipage}
  \hfill
  \begin{minipage}[b]{0.49\textwidth}
    \centering
     \begin{tabular}{c|cc|cc}
 \hline
  \multirow{2}{*}{Methods}&\multicolumn{2}{c|}{Foggy Zurich}&\multicolumn{2}{c}{Foggy Driving}\\
  \cline{2-5}
  &R&B&R&B\\
  \hline
  FC+FS\cite{hahner2019semantic}&41.4&30.9&\textbf{50.7}&35.2\\
  AC+FS&\textbf{43.8}&\textbf{32.9}&\textcolor{red}{50.3}&\textbf{39.9}\\
  \hline
 \end{tabular}
\captionof{table}{\label{mix_fggan} Results of semantic segmentation on the Foggy Zurich and Foggy Driving dataset based on RefineNet (R) with ResNet-101 backbone and BiseNet (B) with ResNet-18 backbone using different simulated foggy images. The results are reported on mIoU over 19 categories. The best result is denoted in bold. "FC", "FS", "AC" represent "Foggy Cityscapes", "Foggy Synscapes", "AnalogicalGAN Cityscapes", respectively.}
    \end{minipage}
  \end{minipage}

\subsection{Semantic Foggy Scene Understanding}
\subsubsection{Experiments Setup}
In this section, we validate the usefulness of our translated images for the downstream task semantic foggy scene understanding. Specifically, following the paradigm in \cite{sakaridis2018semantic, hahner2019semantic}, the pretrained semantic segmentation model on the real clear weather images, Cityscapes, is fine-tuned on the synthesized foggy images. Then the fine-tuned model is tested on two real foggy image datasets: Foggy Zurich\cite{dai2019curriculum} and Foggy Driving \cite{sakaridis2018semantic}. We compare the semantic foggy scene understanding performance of our AnalogicalGAN model translation results with the state-of-the-art physics-based foggy image synthesis results, Foggy Cityscapes\cite{sakaridis2018semantic}, and the translation results of the traditional image translation methods CycleGAN and MUNIT as shown in Section \ref{sec:inner_trans}. In addition to the setting \emph{Virtual KITTI to Cityscapes} as used in Section \ref{sec:inner_trans}, we further evaluate all methods in another setting \emph{Virtual KITTI to Synscapes}. The performance of foggy scene understanding of all methods are reported for both of the two translation settings.  

\textbf{Synscapes} is a synthetic dataset consisting of 25,000 clear weather images imitating the content and structure of Cityscapes dataset. Pixel-wise ground-truth semantic labels and depth maps are given in the dataest.

\textbf{Foggy Zurich} consists of 3,808 foggy scene images taken from Zurich City, 40 of which are densely labeled. We use them as test data in our experiment.

\textbf{Foggy Driving} is a dataset containing 101 real foggy images collected in various areas of Zurich and from the Internet. The dataset is annotated coarsely and the classes are compatible with Cityscapes dataset.

As shown in \cite{dai2019curriculum}, the fog density of the synthesized foggy image highly affects the semantic foggy scene understanding performance. Our AnalogicalGAN model can control the density of the synthesized fog via the domainness variable $z$.
In order to generate the foggy image with the appropriate fog density, during testing stage, the domainness variable $z$ is set to 0.88 and 0.9 for Cityscapes and Synscapes, respectively. For semantic segmentation, we follow the paradigm and fine-tuning details in \cite{sakaridis2018semantic} and \cite{hahner2019semantic}. The RefineNet \cite{lin2017refinenet} with ResNet-101 backbone \cite{he2016deep} and the BiseNet \cite{yu2018bisenet} with ResNet-18 backbone \cite{he2016deep} are utilized as the semantic segmentation networks. 
\begin{table}
\begin{subtable}{0.49\textwidth}
\begin{tabular}{c|cc|cc}
 \hline
 \multicolumn{5}{c}{Virtual KITTI$\rightarrow$ Cityscapes}\\
 \hline
  \multirow{3}{*}{Fine-tuning}& \multicolumn{4}{c}{Testing}\\
  \cline{2-5}
  &\multicolumn{2}{c|}{FZ}&\multicolumn{2}{c}{FD}\\
  \cline{2-5}
  &R&B&R&B\\
  \hline
  Cityscapes\cite{hahner2019semantic}&34.6&16.1&44.3&27.2\\
  FC\cite{hahner2019semantic}&36.9&25.0&46.1&30.3\\
  CycleGAN\cite{zhu2017unpaired}&40.5&27.1&47.7&30.0\\
  MUNIT\cite{huang2018munit}&39.1&26.0&\textbf{47.8}&30.5\\
  AC(ours)&\textbf{42.3}&\textbf{28.4}&\textcolor{red}{47.5}&\textbf{30.8}\\
  \hline
 \end{tabular}
\caption{\label{table_cityscapes}}
\end{subtable}
\begin{subtable}{0.49\textwidth}
\begin{tabular}{c|cc|cc}
 \hline
 \multicolumn{5}{c}{Virtual KITTI$\rightarrow$ Synscapes}\\
 \hline
  \multirow{3}{*}{Fine-tuning}& \multicolumn{4}{c}{Testing}\\
  \cline{2-5}
  &\multicolumn{2}{c|}{FZ}&\multicolumn{2}{c}{FD}\\
  \cline{2-5}
  &R&B&R&B\\
  \hline
  Cityscapes\cite{hahner2019semantic}&34.6&16.1&44.3&27.2\\
  FS\cite{hahner2019semantic}&40.3&27.8&48.4&30.9\\
  CycleGAN\cite{zhu2017unpaired}&41.6&30.9&47.8&33.1\\
  MUNIT\cite{huang2018munit}&40.5&27.5&48.3&32.8\\
  AS(ours)&\textbf{41.8}&\textbf{31.5}&\textbf{49.8}&\textbf{34.2}\\
  \hline
 \end{tabular}
\caption{\label{table_synscapes}}
\end{subtable}
\caption{Results of semantic segmentation on the Foggy Zurich and Foggy Driving dataset based on RefineNet (R) with ResNet-101 backbone and BiseNet (B) with ResNet-18 backbone using different simulated foggy images. The results are reported on mIoU over 19 categories. The best result is denoted in bold. "FC", "AC", "AS", "FD", "FZ" represent "Foggy Cityscapes", "AnalogicalGAN Cityscapes", "AnalogicalGAN Synscapes", "Foggy Driving", "Foggy Zurich", respectively.}
\end{table}

\subsubsection{Experiments Results}
The results of semantic foggy scene understanding based on the synthesized foggy images from Cityscapes and Synscapes are shown in Table \ref{table_cityscapes} and Table \ref{table_synscapes}, respectively. 
In Table \ref{table_cityscapes} and Table \ref{table_synscapes}, while using Cityscapes and Synscapes as real clear weather images, it is shown that our AnalogicalGAN outperforms the physics-based foggy image synthesis methods "Foggy Cityscapes" and "Foggy Synscapes". The improvement is consistent on both Foggy Zurich and Foggy Driving, and with RefineNet and with BiseNet segmentatin networks. When compared to the traditional image translation methods, our "AnalogicalGAN" outperforms both "CycleGAN" and "MUNIT" on both test sets and for both segmentation networks, except for one case (when utilizing the RefineNet and testing on Foggy Driving) in which our method reaches comparable performance with MUNIT ($47.5\%$ v.s. $47.8\%$). 

Moreover, following \cite{hahner2019semantic}, by mixing the "Foggy Synscapes" with "AnalogicalGAN Cityscapes", i.e. Cityscapes translated with "AnalogicalGAN" model, the performance can be further improved. From Table \ref{mix_fggan}, it is shown that the mixture of "AnalogicalGAN Cityscapes" and "Foggy Synscapes" improves the performance of the state-of-the-art methods, mixture of "Foggy Citysacpes" and "Foggy Synscapes" by $2.4\%$ and $2.0\%$ on Foggy Zurich with RefineNet and BiseNet, while improving by $4.7\%$ on Foggy Driving with BiseNet and reaching comparable performance, 50.3$\%$ v.s. 50.7$\%$, on Foggy Driving with RefineNet. The semantic foggy scene understanding performance and comparison demonstrate the effectiveness of our AnalogicalGAN model for synthesizing fog effects to real images.  The results also shows the advantage of our proposed method over the physics-based fog synthesis methods and the traditional image translation methods. More detailed results on each classes are listed in the supplementary material due to space limitation.

\section{Conclusion}
\label{sec:conclusion}
In this work, we have presented AnalogicalGAN, a novel analogical image translation framework. Different from the traditional image translation,  analogical image translation is able to achieve the zero-shot image translation capability via analogy. Applying our AnalogicalGAN model to the fog generation task in which the synthetic clear-weather images, synthetic foggy images, and the real clear-weather images are given, our AnalogicalGAN model is able to synthesize realistic fog effects into real clear-weather images, even though no real foggy images is available in both the training and testing stages. The qualitative and quantitative comparison and the evaluation on semantic foggy scene understanding  prove the effectiveness of our AnalogicalGAN model. Extending our AnalogicalGAN model to other analogical image translation scenarios constitutes our future work. 
\section*{Broader Impact}
In this paper, we propose the "AnalogicalGAN" model, a kind of analogical image translation framework. It can be seen as the zero-shot generalization of existing image-to-image translation framework. 

The analogical image translation framework has the potential to highly reduce the gathering and labeling difficulty of the data. Benefiting from the transferred data scale and diversity, the deep model is expected to be more robust, reliable and effective under different even extreme conditions, which is able to promote and accelerate the launch of deep-based system such as the medical computer-assisted system and autonomous driving system. 

The easy availability of the transferred labeled data and the launch of the more reliable and effective deep-based systems likely have complex social impacts. (i) On one hand, transferred labeled data will save much cost on the data gathering and labeling and avoid the wasteful duplication of labor. More and more deep-based artificial intelligent systems will become part of the people's life, bringing convenience, wealth and prosperity. (ii) On the other hand, the transferred labeled data might induce the unemployment for the people who are engaged in gathering and labeling the dataset. Meanwhile, the launch of artificial intelligent systems may also cause the job loss. Besides, another concern is that the techniques for synthesizing the image is possible to be used for the illegal purpose of forgery and deception. 

We would encourage further work on the detection of the forgery and deception of the image even though the detection will become harder and harder as the image synthesis techniques develop. From the view of long-term development, in order to mitigate the risks of image synthesis, more regulations and guidance on tracking and stopping the harmful and dangerous synthesized images should be made. 

{\small
\bibliographystyle{ieee}
\bibliography{egbib}
}
\pagebreak

\setcounter{equation}{0}
\setcounter{figure}{0}
\setcounter{table}{0}
\setcounter{page}{1}
\setcounter{section}{0}

\renewcommand{\thesection}{S\arabic{section}}  
\renewcommand{\thefigure}{S\arabic{figure}}
\renewcommand{\thetable}{S\arabic{table}}
\section*{Supplementary}
In this supplementary material, we provide the additional information for,
\begin{itemize}
    \item[\textbf{S1}] detailed architecture and implementation for AnalogicalGAN model,
    \item[\textbf{S2}] ablation study of our AnalogicalGAN model,
    \item[\textbf{S3}] comparison of our AnalogicalGAN model with baseline method encoding auxiliary information,
    \item[\textbf{S4}] more visual results for fog generation on Cityscapes and Synscapes,
    \item[\textbf{S5}] more detailed quantitative results on semantic foggy scene understanding. 
\end{itemize}
\section{Architecture and Implementation of AnalogicalGAN Model}
\begin{figure}[t]
\centering
\includegraphics[width=1.0\linewidth]{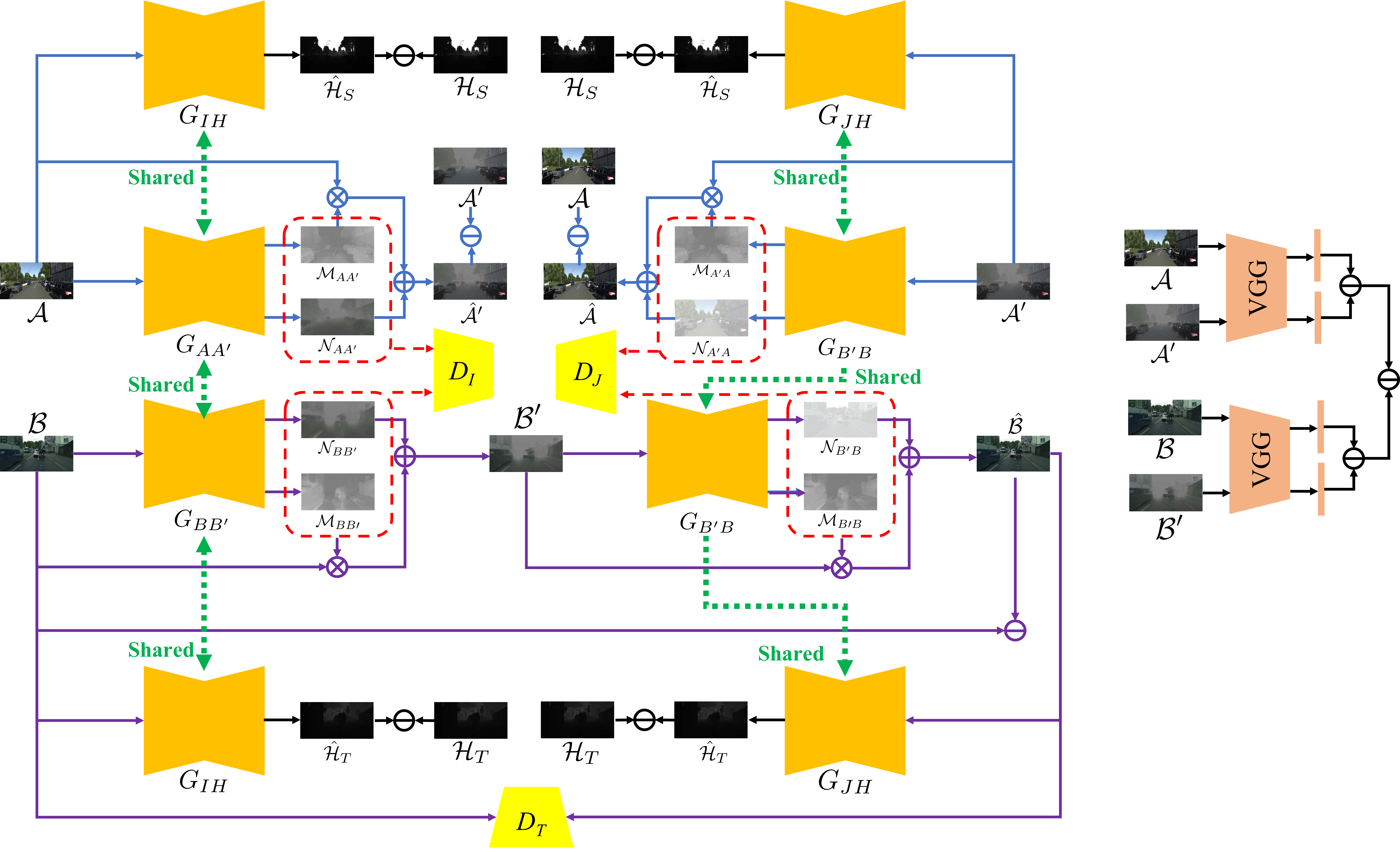}
\caption{Network architecture visualization of our AnalogicalGAN model. Due to paired samples $\cA$ and $\cA\prime$ on the source domain available, the generators $G_{AA^\prime}$ and $G_{A^\prime A}$ are trained in the supervised way on the source domain, shown with blue arrows. And the \emph{gist} is measured with the alignment map $\cM_{AA^\prime}$, $\cM_{A^\prime A}$ and residual map $\cN_{AA^\prime}$, $\cN_{A^\prime A}$. Then the \emph{gist}, $\cM_{AA^\prime}$, $\cM_{A^\prime A}, \cN_{AA^\prime}$, $\cN_{A^\prime A}$, is transferred to the target domain, $\cM_{BB^\prime}$, $\cM_{B^\prime B}, \cN_{BB^\prime}$, $\cN_{B^\prime B}$, through the adversarial learning, shown with red dash arrows and boxes. Moreover, the cycle-consistency on the target domain is utilized to guarantee that the \emph{gist} is compatible with the target domain, shown with purple arrows. Due to the high correlation between the fog and depth, the depth map is leveraged as the auxiliary information, by sharing parameters between the depth generators $G_{IH}, G_{JH}$ and the translation generators $G_{AA^\prime}$, $G_{A^\prime A}$, $G_{BB^\prime}$ and $G_{B^\prime B}$. Besides, the perceptual loss based on the VGG feature distance is introduced to assist the image translation process. The auxiliary depth information and the perceptual loss are shown with black arrows. In total, the blue, purple, green, red and black arrows is corresponding to the supervised module, the cycle consistent module, the adversarial module and the auxiliary module, respectively.}
\label{supfig:network}
\end{figure}

In Section 3.2 of the main paper, we introduce that our proposed AnalogicalGAN model is composed of four modules, the supervised module, the adversarial module, the cycle consistent module and the auxiliary module. In detail, we introduce the architecture and the implementation of our AnalogicalGAN model here. In Fig. \ref{supfig:network}, the detailed network architecture is shown. Consistent with Fig. 2 in the main paper, the blue, purple, red and black arrows represent the supervised module, the cycle consistent module, the adversarial module and the auxiliary module, respectively. The generators and discriminators are implemented with the generators and discriminators network structure of CycleGAN \cite{zhu2017unpaired}. Besides, The generator $G_{AA^\prime}$ and $G_{BB^\prime}$, $G_{A^\prime A}$ and $G_{B^\prime B}$ share all the parameters, respectively. The translation generator $G_{AA^\prime}, G_{BB^\prime}$ and depth generator $G_{IH}$ share all the parameters except for the final deconvolution layer, and $G_{A^\prime A}, G_{B^\prime B}$ and $G_{JH}$ acts in the similar way. In addition, following \cite{huang2018munit}, we adopt the VGG feature (relu4$\_$3) to compute the perceptual loss. 

\section{Ablation Study of AnalogicalGAN model}
Our AnalogicalGAN model consists of the supervised module, the adversarial module, the cycle consistent module and the auxiliary module. And there are the depth loss and the perceptual loss covered in the auxiliary module. In this section, we show the qualitative and quantitative ablation study results of the full objective proposed in Section 3.2 of the main paper, and analyze the effect of different modules and loss terms to prove that each of them are effective for our analogical image translation (AIT) task. 

\subsection{Qualitative Ablation Study Results}
\begin{figure}
    \centering
    \includegraphics[width=1.0\textwidth]{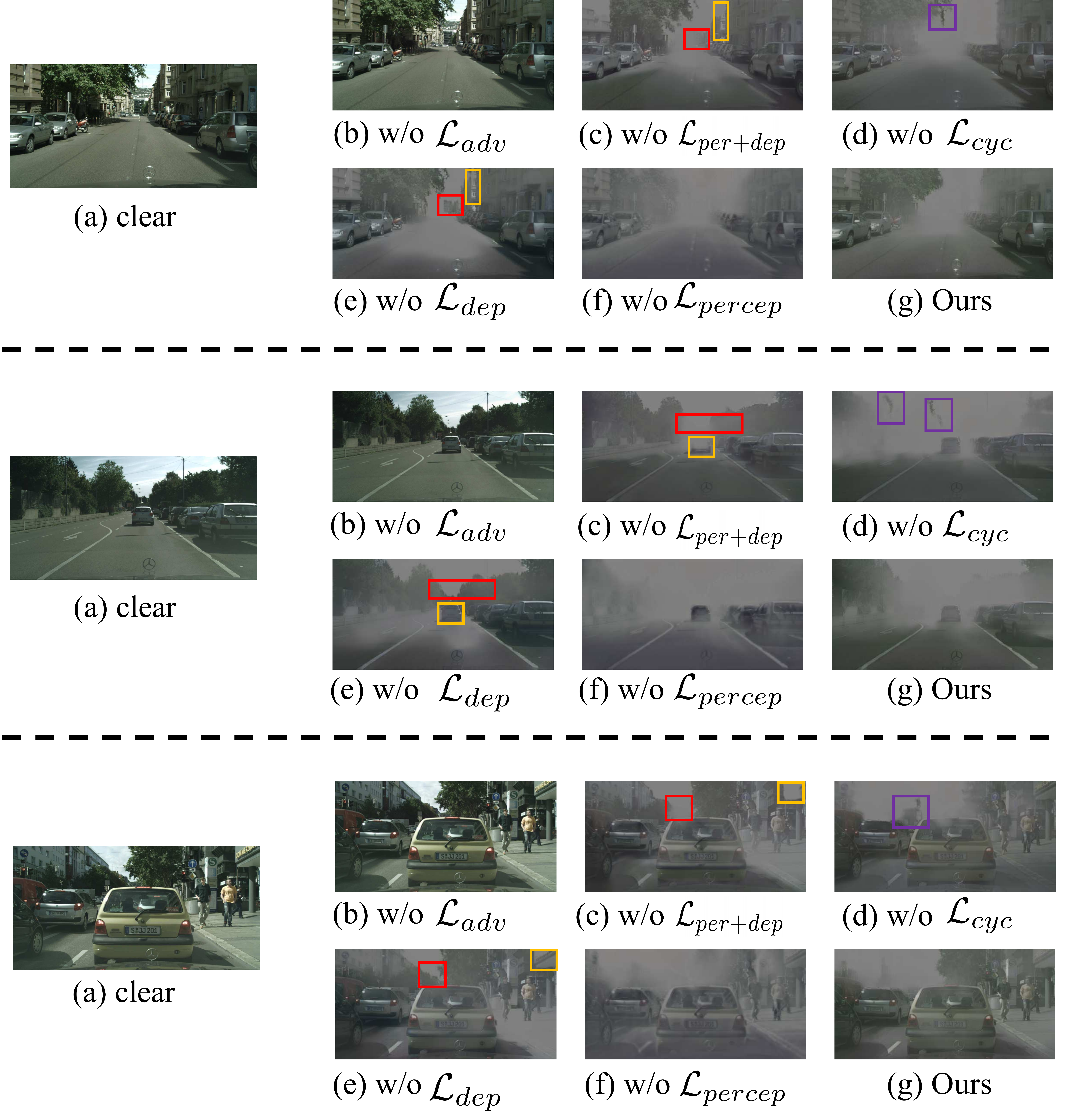}
    \caption{Qualitative comparison against the ablations of the full objective. Without the adversarial module $\cL_{adv}$ (in (b)), the fog effect cannot be generated. Without the cycle consistency module $\cL_{cyc}$ (in (d)), there are inconsistent fog effect artifacts shown, which is labeled with purple box. Without the auxiliary module $\cL_{per+dep}$, i.e. $\cL_{percep}$ and $\cL_{depth}$ (in (c)), the distance-wise feature of the fog is not well generated and the real feature of the objects is not well preserved such as the middle car in (c) of the second row. Within the auxiliary module, purely adding the perceptual loss $\cL_{percep}$ (in (e)) makes the real feature of objects more obvious (comparison between the orange box part in (c) and (e)) but weakens the fog effect (comparison between the red box part in (c) and (e)). Also, within the auxiliary module, purely adding the depth loss $\cL_{dep}$ (in (f)) strengthens the distance-wise fog feature but causes the loss of the real feature of the objects. By adopting the full objective, as shown in (g), the fog effect is well generated, showing the consistent and distance-wise fog feature and preserving the real feature of the objects.}
    \label{fig:ablation}
\end{figure}

In Fig. \ref{fig:ablation}, we adopt the fog generation task, as done in Section 4.1 of the main paper, to show the qualitative comparison against the ablations of the full objective. From Fig. \ref{fig:ablation}(b), it is shown that the adversarial module is essential for the fog generation, without which the generated image is almost the same as the original real clear weather image in Fig. \ref{fig:ablation}(a). From Fig. \ref{fig:ablation}(d) and Fig. \ref{fig:ablation}(g), it is shown that the cycle consistent module makes the generated fog effect more consistent, i.e. avoid the clear islet in the translated fog part (refer to the purple box in Fig. \ref{fig:ablation}(d)). From Fig. \ref{fig:ablation}(c) and Fig. \ref{fig:ablation}(g), the auxiliary module helps strengthen the distance-wise fog effect and well preserve the real feature of the objects such as the building, the tree and the car. Moreover, due to there are two terms, the perceptual loss $\cL_{percep}$ and the depth loss $\cL_{depth}$ included in the auxiliary module. Further ablations of the perceptual loss and the depth loss are compared against. From Fig. \ref{fig:ablation}(c) and \ref{fig:ablation}(e), purely adding the perceptual loss of the auxiliary module can help preserve the real feature of the objects (refer to orange box part in Fig. \ref{fig:ablation}(c) and \ref{fig:ablation}(e)), but weakens the fog effect of the translated image (refer to red box part in Fig. \ref{fig:ablation}(c) and \ref{fig:ablation}(e)). From Fig. \ref{fig:ablation}(c) and \ref{fig:ablation}(f), purely adding the depth loss of the auxiliary module can enhance the distance-wise feature of the fog effect but cause the loss of the real feature of the objects in some extent such as the the building, the ground and the car objects in Fig. \ref{fig:ablation}(f). As shown in Fig. \ref{fig:ablation}(g), by introducing all the modules and losses in the full objective, the modules can compensate and promote for each other, and generate the consistent and distance-wise fog effects while preserving the real feature of the objects at the same time. 
\begin{table*}[h]
\setlength{\tabcolsep}{3pt}
\centering
 \resizebox{\textwidth}{18mm}
 {
 \begin{tabular}{c|c|c|ccccccccccccccccccc|c}
 \hline
  \rotatebox{90}{Testing}&\rotatebox{90}{Model}&\rotatebox{90}{Fine-Tuning}&\rotatebox{90}{road}&\rotatebox{90}{sidewalk}&\rotatebox{90}{building}&\rotatebox{90}{wall}&\rotatebox{90}{fence}&\rotatebox{90}{pole}&\rotatebox{90}{traffic light}&\rotatebox{90}{traffic sign}&\rotatebox{90}{vegetation}&\rotatebox{90}{terrian}&\rotatebox{90}{sky}&\rotatebox{90}{person}&\rotatebox{90}{rider}&\rotatebox{90}{car}&\rotatebox{90}{truck}&\rotatebox{90}{bus}&\rotatebox{90}{train}&\rotatebox{90}{motorbike}&\rotatebox{90}{bicycle}&mIoU\\
  \hline
  \multirow{4}{*}{\rotatebox{90}{FZ}}&\multirow{2}{*}{R}&FC+FS\cite{hahner2019semantic}&87.5&\textbf{60.6}&46.0&\textbf{41.1}&38.5&\textbf{48.2}&62.4&61.9&67.3&38.1&74.4&6.2&22.5&80.8&0.0&\textbf{1.7}&-&45.9&3.8&41.4\\
  &&AC+FS&\textbf{87.7}&51.4&\textbf{58.5}&32.3&\textbf{43.6}&47.3&\textbf{62.5}&\textbf{62.7}&\textbf{75.3}&\textbf{52.3}&\textbf{89.7}&\textbf{7.0}&\textbf{26.2}&\textbf{81.5}&0.0&0.0&-&\textbf{46.7}&\textbf{8.1}&\textbf{43.8}\\
  \cline{2-23}
  &\multirow{2}{*}{B}&FC+FS\cite{hahner2019semantic}&\textbf{81.1}&41.5&\textbf{60.3}&\textbf{33.5}&28.5&21.8&34.4&40.5&\textbf{68.0}&48.2&\textbf{87.9}&0.2&1.1&39.0&0.0&\textbf{0.1}&-&0.0&0.0&30.9\\
  &&AC+FS&60.3&\textbf{43.0}&28.5&22.1&\textbf{30.5}&\textbf{35.5}&\textbf{57.3}&\textbf{59.5}&56.4&\textbf{53.1}&56.5&\textbf{7.4}&\textbf{24.5}&\textbf{60.4}&0.0&0.0&-&\textbf{24.6}&\textbf{6.1}&\textbf{32.9}\\
  \hline
  \multirow{4}{*}{\rotatebox{90}{FD}}&\multirow{2}{*}{R}&FC+FS\cite{hahner2019semantic}&\textbf{92.4}&\textbf{34.0}&\textbf{76.1}&\textbf{23.9}&16.2&\textbf{45.6}&\textbf{55.9}&\textbf{61.6}&\textbf{76.4}&\textbf{11.1}&92.2&57.5&45.6&69.9&13.7&42.3&\textbf{82.2}&14.1&52.6&\textbf{50.7}\\
  &&AC+FS&91.4&22.5&76.0&18.2&\textbf{21.3}&41.3&55.6&58.7&75.6&\textbf{11.1}&\textbf{93.8}&\textbf{58.8}&\textbf{46.2}&\textbf{73.9}&\textbf{18.6}&\textbf{47.0}&70.5&\textbf{17.6}&\textbf{57.1}&\textcolor{red}{50.3}\\
  \cline{2-23}
  &\multirow{2}{*}{B}&FC+FS\cite{hahner2019semantic}&\textbf{84.3}&\textbf{23.8}&\textbf{68.0}&4.0&\textbf{7.3}&29.6&39.4&45.7&66.4&3.7&89.7&36.1&6.0&62.7&10.0&\textbf{37.7}&18.7&0.0&35.5&35.2\\
  &&AC+FS&80.4&20.5&65.5&\textbf{9.3}&6.2&\textbf{36.6}&\textbf{52.7}&\textbf{50.0}&\textbf{67.0}&\textbf{9.9}&\textbf{93.1}&\textbf{48.6}&\textbf{18.4}&\textbf{62.8}&\textbf{12.5}&26.3&\textbf{25.4}&\textbf{21.2}&\textbf{52.3}&\textbf{39.9}\\
  \hline
 \end{tabular}
}
\caption{\label{sup:mix_fggan} Results of semantic segmentation on the Foggy Zurich (FZ) and Foggy Driving dataset (FD) based on RefineNet model (R) with ResNet-101 backbone and BiseNet (B) with ResNet-18 backbone using different simulated foggy images. The results are reported on mIoU over 19 categories. The best result is denoted in bold. "FC", "FS", "AC" represent "Foggy Cityscapes", "Foggy Synscapes", "AnalogicalGAN Cityscapes", respectively.}
\end{table*}

\subsection{Quantitative Ablation Study Results}
In order to further explore the effect of the different loss terms in the full objective quantitatively, we compare our model with the ablations of the full objective for the semantic foggy scene understanding, as done in Section 4.2 of the main paper. In Table. \ref{table_ablation}, we show the semantic foggy scene understanding performance of our AnalogicalGAN model and the ablations of the full objective. The RefineNet with ResNet-101 backbone is adopted as the semantic segmentation network and the fine-tuned model is tested on the Foggy Zurich \cite{dai2019curriculum} dataset. From Table. \ref{table_ablation}, it is shown that our AnalogicalGAN model with full objective outperforms the ablations of the full objective. Also, purely adding the perceptual loss as the auxiliary module, i.e., w/o $\cL_{dep}$ in Table. \ref{table_ablation}, causes the performance drop compared with the one without auxiliary module, i.e. w/o $\cL_{per+dep}$ in Table. \ref{table_ablation}, $40.8\%$ v.s. $41.7\%$. It is due to that purely adding the perceptual loss weakens the fog effect as we analyze in the qualitative ablation study part. 

\begin{table*}[h]
\setlength{\tabcolsep}{3pt}
\centering
 \resizebox{\textwidth}{18mm}
 {
 \begin{tabular}{c|ccccccccccccccccccc|c}
 \hline
  \rotatebox{90}{Fine-Tuning}&\rotatebox{90}{road}&\rotatebox{90}{sidewalk}&\rotatebox{90}{building}&\rotatebox{90}{wall}&\rotatebox{90}{fence}&\rotatebox{90}{pole}&\rotatebox{90}{traffic light}&\rotatebox{90}{traffic sign}&\rotatebox{90}{vegetation}&\rotatebox{90}{terrian}&\rotatebox{90}{sky}&\rotatebox{90}{person}&\rotatebox{90}{rider}&\rotatebox{90}{car}&\rotatebox{90}{truck}&\rotatebox{90}{bus}&\rotatebox{90}{train}&\rotatebox{90}{motorbike}&\rotatebox{90}{bicycle}&mIoU\\
  \hline
  Cityscapes\cite{lin2017refinenet}&74.3&\textbf{56.5}&35.5&20.2&23.8&39.6&54.4&58.3&58.3&28.9&66.8&1.6&27.4&\textbf{81.7}&0.0&0.0&-&21.1&6.2&34.6\\
  w/o $L_{adv}$&63.2&49.2&47.0&14.1&23.8&33.9&46.0&56.3&45.4&23.7&78.9&0.3&21.1&80.5&0.0&0.0&-&32.8&7.1&32.8\\
  w/o $L_{cyc}$&87.3&51.2&48.0&27.2&28.4&46.8&61.7&57.1&70.6&41.4&82.6&4.7&\textbf{31.5}&79.0&0.0&36.1&-&40.5&4.3&42.0\\
  w/o $L_{per+dep}$&\textbf{89.4}&51.7&44.3&\textbf{35.1}&32.1&48.2&60.6&60.0&74.7&44.8&78.3&4.8&25.7&80.9&0.0&12.4&-&45.9&3.8&41.7\\
  w/o $L_{percep}$&88.2&49.2&\textbf{58.3}&29.7&\textbf{36.8}&47.3&\textbf{61.8}&50.4&74.0&40.0&\textbf{89.2}&1.8&11.4&80.4&0.0&\textbf{48.4}&-&20.2&8.8&41.9\\
  w/o $L_{dep}$&83.1&56.4&34.3&18.0&31.8&46.0&60.1&59.9&66.6&\textbf{48.1}&65.3&5.6&27.9&81.5&0.0&38.9&-&\textbf{47.6}&3.9&40.8\\
  Ours&88.1&55.8&43.0&29.3&33.2&\textbf{50.4}&61.6&\textbf{60.5}&\textbf{75.3}&43.8&75.8&\textbf{6.5}&28.7&80.3&0.0&5.1&-&46.8&\textbf{20.2}&\textbf{42.3}\\
  \hline
 \end{tabular}
}
\caption{\label{table_ablation} Ablation study results of semantic segmentation on the Foggy Zurich dataset based on RefineNet model with ResNet-101 backbone. The results are reported on mIoU over 19 categories. The best result is denoted in bold. It is shown that our AnalogicalGAN model with the full objective achieves the highest performance compared with the variants of the model which exclude different terms in the full objective.}
\end{table*}

\section{Comparison with Baseline Method Encoding Auxiliary Information}
In Section 4.1 and 4.2 of the main paper, we adopt the traditional image-to-image translation model CycleGAN and MUNIT as the baseline method to show the advantage of our AnalogicalGAN model on the AIT task. In our AnalogicalGAN model shown in Fig. \ref{supfig:network}, the auxiliary depth information is encoded into the translation generator $G_{BB^\prime}, G_{B^\prime B}$, by sharing the parameters between the depth generator $G_{IH}, G_{JH}$ and the translation generator $G_{BB^\prime}, G_{B^\prime B}$, respectively. In this way, we can adopt the same strategy to encode the depth information into CycleGAN to compare our AnalogicalGAN model with the traditional image translation model encoding the depth information. Due to the auto-encoder structure of the MUNIT model for encoding and decoding the style code and content code, it is not suitable for encoding the depth information into MUNIT model with our strategy and we only take CycleGAN for encoding depth information and serving as comparison in this section.

\begin{figure}
    \centering
    \includegraphics[width=1.0\textwidth]{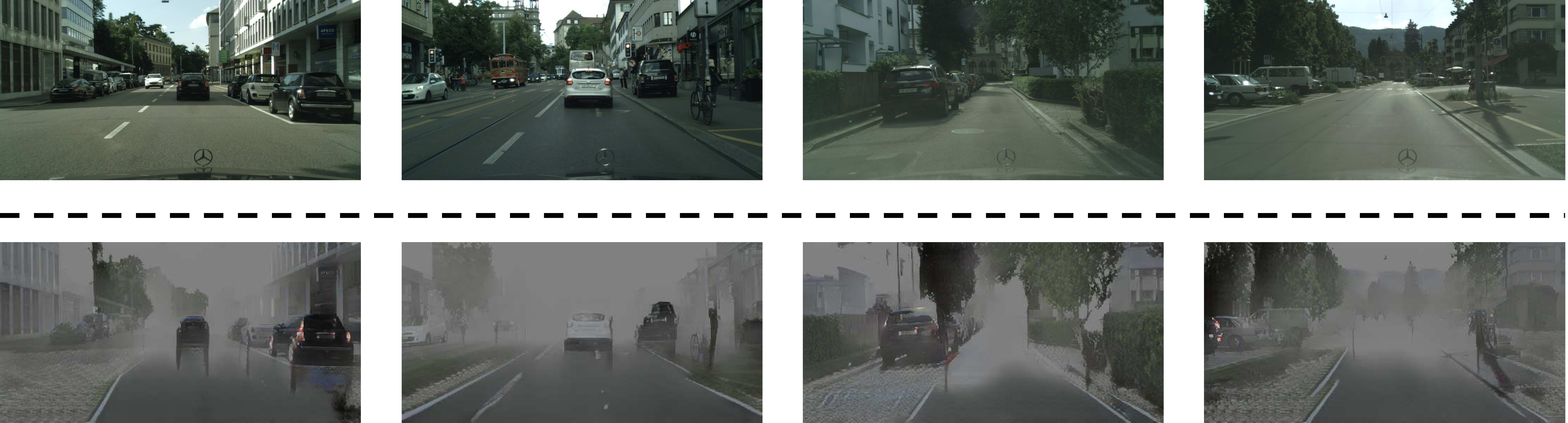}
    \caption{Qualitative fog generation results of the CycleGAN model encoding the depth information. The depth information of $\cB$ is encoded by sharing the parameters between translation generator and depth generator. The model is trained to translate between $\cB$ and $\cA^\prime$ and tested on $\cB$, which is different from the main paper where the CycleGAN model is trained to translate between $\cA$ and $\cA^\prime$ and tested on $\cB$. The latter training and testing paradigm for CycleGAN model , which is adopted by the main paper, can help weaken the synthetic feature inherited from the synthetic foggy weather image. However, in order to encode the depth information of $\cB$ into the CycleGAN model, we can only adopt the former training and testing paradigm. The generated real foggy weather image with CycleGAN model encoding depth information highly suffers from the synthetic feature artifacts, which is inherited from the synthetic foggy weather image.}
    \label{subfig:cyclegan_depth}
\end{figure}

In Section 4.1 and 4.2 of the main paper, the CycleGAN model is trained to translate between $\cA$ and $\cA^\prime$ and tested on $\cB$ to generate $\cB^\prime$, while the MUNIT model is trained to translate between $\cB$ and $\cA^\prime$ and tested on $\cB$ to generate $\cB^\prime$. Adopting this paradigm, the CycleGAN and MUNIT model can generate their own best translation result respectively under our AIT task setting. However, in this section, in order to encode the depth information of $\cB$ into CycleGAN model, the paradigm for trainging and testing CycleGAN model needs to be changed, i.e. the CycleGAN model is trained to translate between $\cB$ and $\cA^\prime$ and tested on $\cB$ to generate $\cB^\prime$. Then the depth information of $\cB$ is encoded by sharing the parameters between the depth generator and the translation generator as we discuss above. In Fig. \ref{subfig:cyclegan_depth} and Table \ref{table_cyclegan_depth}, we show the qualitative and quantitative results of the CycleGAN model encoding the depth information, respectively. From Fig. \ref{subfig:cyclegan_depth}, it is shown that the generated real foggy weather image has obvious artifacts due to the synthetic feature inherited from the synthetic foggy weather image. As shown in Table \ref{table_cyclegan_depth}, the synthetic artifacts also cause the extreme semantic foggy scene understanding performance drop even compared with the pretrained model on Cityscapes, $28.1\%$ v.s. $34.6\%$.
\begin{table*}[h]
\setlength{\tabcolsep}{3pt}
\centering
 \resizebox{\textwidth}{11mm}
 {
 \begin{tabular}{c|ccccccccccccccccccc|c}
 \hline
  \rotatebox{90}{Fine-Tuning}&\rotatebox{90}{road}&\rotatebox{90}{sidewalk}&\rotatebox{90}{building}&\rotatebox{90}{wall}&\rotatebox{90}{fence}&\rotatebox{90}{pole}&\rotatebox{90}{traffic light}&\rotatebox{90}{traffic sign}&\rotatebox{90}{vegetation}&\rotatebox{90}{terrian}&\rotatebox{90}{sky}&\rotatebox{90}{person}&\rotatebox{90}{rider}&\rotatebox{90}{car}&\rotatebox{90}{truck}&\rotatebox{90}{bus}&\rotatebox{90}{train}&\rotatebox{90}{motorbike}&\rotatebox{90}{bicycle}&mIoU\\
  \hline
  Cityscapes\cite{lin2017refinenet}&74.3&\textbf{56.5}&\textbf{35.5}&20.2&23.8&\textbf{39.6}&54.4&\textbf{58.3}&58.3&28.9&\textbf{66.8}&1.6&\textbf{27.4}&\textbf{81.7}&0.0&0.0&-&\textbf{21.1}&6.2&\textbf{34.6}\\
  CycleGAN w/ depth&\textbf{85.5}&34.6&19.9&\textbf{27.1}&\textbf{24.8}&34.0&\textbf{56.4}&54.5&\textbf{60.7}&\textbf{49.4}&10.0&\textbf{2.7}&0.0&65.8&0.0&0.0&-&0.0&\textbf{7.9}&28.1\\
  \hline
 \end{tabular}
}
\caption{\label{table_cyclegan_depth} Results of semantic segmentation on the Foggy Zurich dataset based on RefineNet model with ResNet-101 backbone. The results are reported on mIoU over 19 categories. The best result is denoted in bold. }
\end{table*}

\section{More Visual Results for Fog Generation}
\begin{figure}
    \centering
    \includegraphics[width=1.0\textwidth]{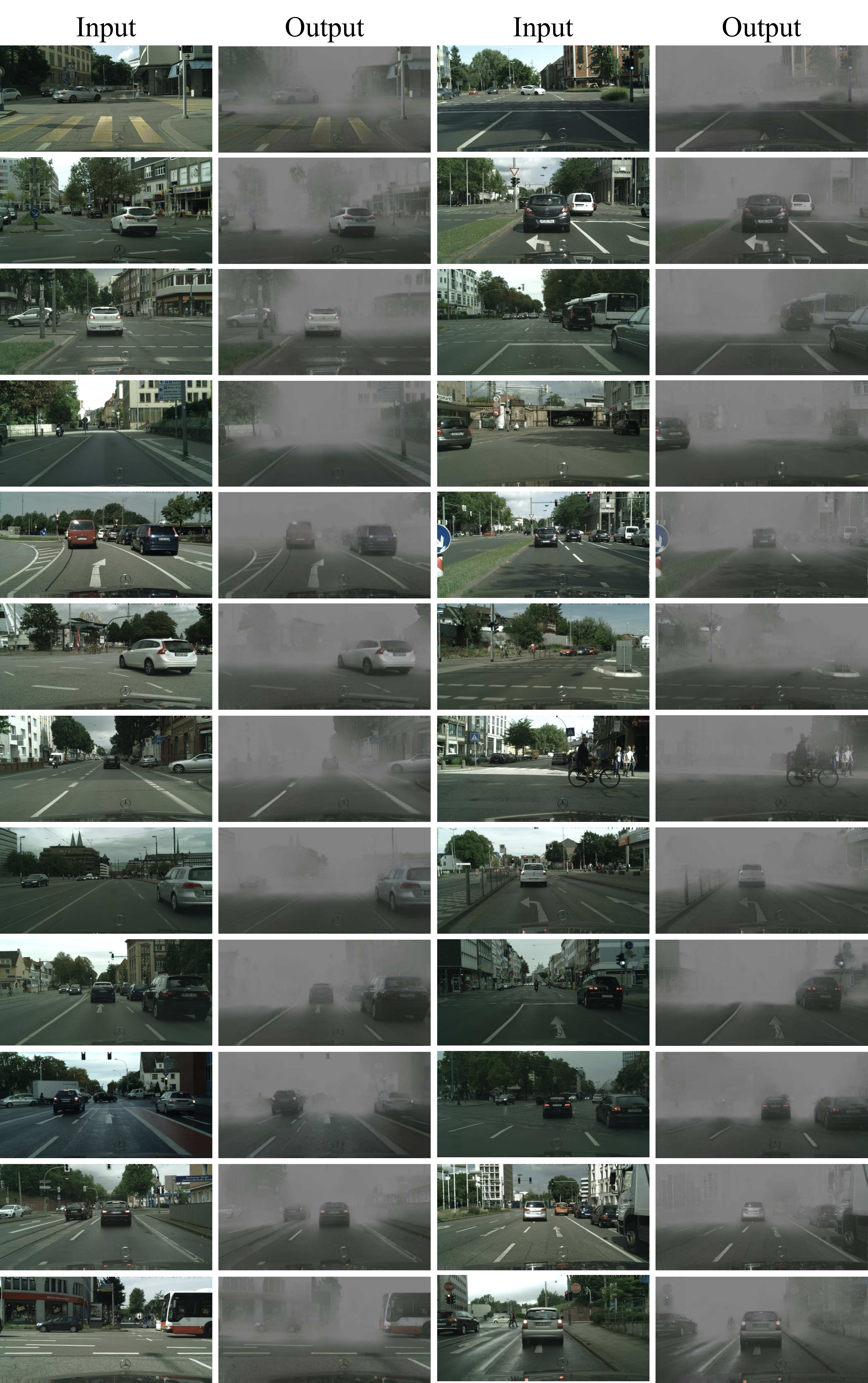}
    \caption{Qualitative results of fog generation on Cityscapes with our AnalogicalGAN model. The input is the real clear weather image from Cityscapes while the output is our translated real foggy weather image.}
    \label{subfig:cityscapes_visul}
\end{figure}

\begin{figure}
    \centering
    \includegraphics[width=1.0\textwidth]{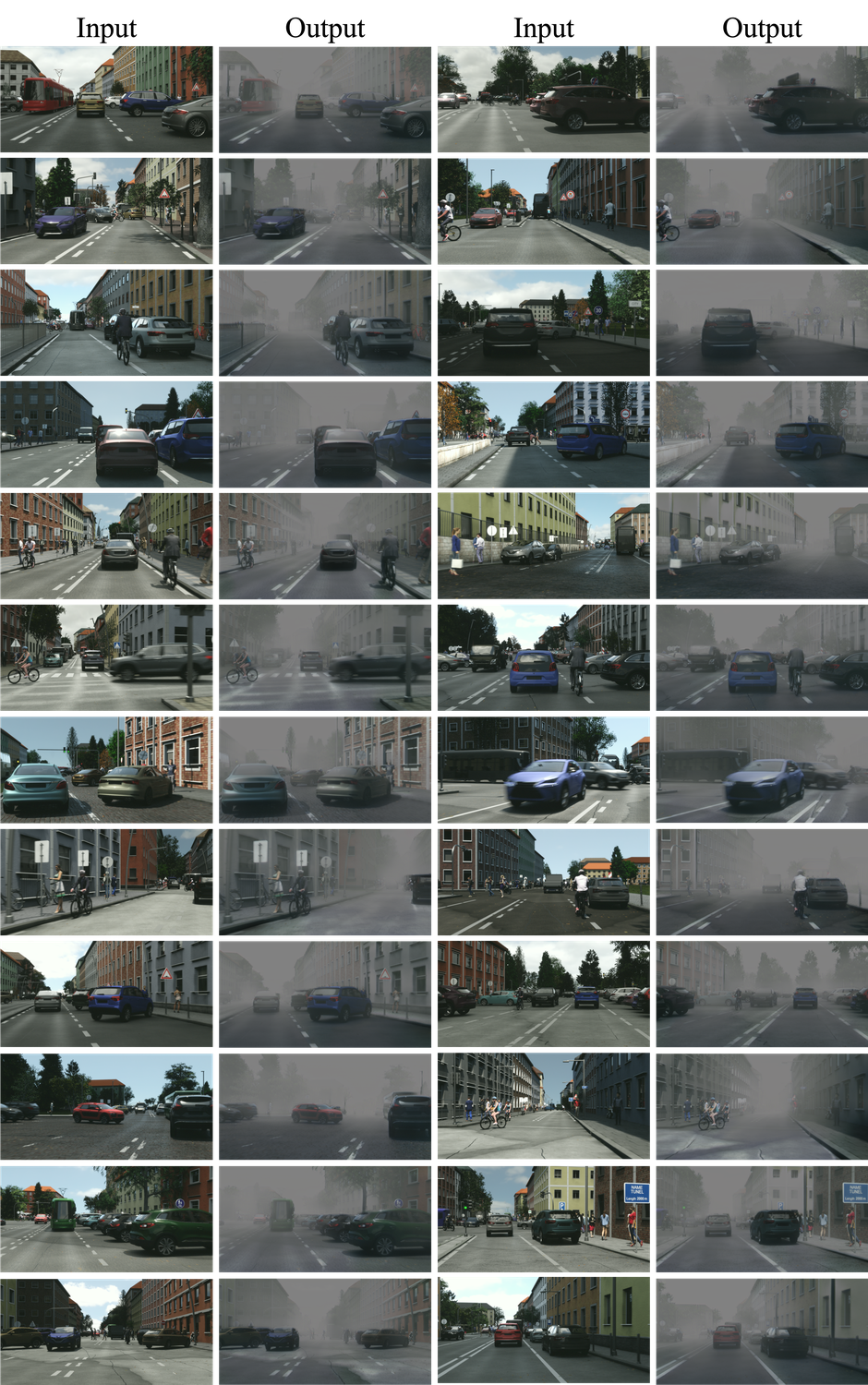}
    \caption{Qualitative results of fog generation on Synscapes with our AnalogicalGAN model. The input is the real clear weather image from Synscapes while the output is our translated real foggy weather image.}
    \label{subfig:synscapes_visul}
\end{figure}

In Fig. 3 of the main paper, we provide the qualitative results of fog generation on Cityscapes~\cite{cordts2016cityscapes} with our AnalogicalGAN model. Here we provide more qualitative results of fog generation based on Cityscapes and Synscapes \cite{wrenninge2018synscapes} in Fig. \ref{subfig:cityscapes_visul} and Fig. \ref{subfig:synscapes_visul}, respectively. It is observed that our AnalogicalGAN model can generate the real foggy weather image based on both of the Cityscapes and Synscapes, which further proves the effectiveness of our AnalogicalGAN model for the AIT task.
\section{Detailed Quantitative Results on Semantic Foggy Scene Understanding}
In Table 1, Table 2 and Table3 of the main paper, we compare the semantic foggy scene understanding performance of our AnalogicalGAN model with that of physics-based foggy image synthesis methods, "Foggy Cityscapes", "Foggy Synscapes", and the traditional image translation methods, "CycleGAN", "MUNIT". Here we provide more detailed results on each class in Table \ref{sup:mix_fggan}, Table \ref{sup:table_cityscapes} and Table \ref{sup:table_synscapes}, corresponding to Table 1, Table 2 and Table 3 of the main paper.
\begin{table*}[h]
\setlength{\tabcolsep}{3pt}
\centering
 \resizebox{\textwidth}{34mm}
 {
 \begin{tabular}{c|c|c|ccccccccccccccccccc|c}
 \hline
 \multicolumn{23}{c}{Virtual KITTI$\rightarrow$Cityscapes}\\
  \hline
  \rotatebox{90}{Testing}&\rotatebox{90}{Model}&\rotatebox{90}{Fine-Tuning}&\rotatebox{90}{road}&\rotatebox{90}{sidewalk}&\rotatebox{90}{building}&\rotatebox{90}{wall}&\rotatebox{90}{fence}&\rotatebox{90}{pole}&\rotatebox{90}{traffic light}&\rotatebox{90}{traffic sign}&\rotatebox{90}{vegetation}&\rotatebox{90}{terrian}&\rotatebox{90}{sky}&\rotatebox{90}{person}&\rotatebox{90}{rider}&\rotatebox{90}{car}&\rotatebox{90}{truck}&\rotatebox{90}{bus}&\rotatebox{90}{train}&\rotatebox{90}{motorbike}&\rotatebox{90}{bicycle}&mIoU\\
  \hline
  \multirow{10}{*}{\rotatebox{90}{Foggy Zurich}}&\multirow{5}{*}{\rotatebox{90}{RefineNet}}&Cityscapes\cite{lin2017refinenet}&74.3&56.5&35.5&20.2&23.8&39.6&54.4&58.3&58.3&28.9&66.8&1.6&27.4&81.7&0.0&0.0&-&21.1&6.2&34.6\\
  &&Foggy Cityscapes\cite{sakaridis2018semantic}&81.2&56.7&36.5&27.5&24.6&44.2&59.6&57.8&48.2&33.6&50.2&5.3&25.3&81.9&0.0&\textbf{29.2}&-&36.0&3.1&36.9\\
  &&CycleGAN\cite{zhu2017unpaired}&83.1&\textbf{57.6}&\textbf{54.3}&28.9&\textbf{36.0}&41.8&59.1&\textbf{63.3}&68.8&40.8&\textbf{85.7}&5.1&\textbf{32.2}&\textbf{85.2}&0.0&0.0&-&19.6&8.4&40.5\\
  &&MUNIT\cite{huang2018munit}&79.4&54.7&48.7&26.8&29.9&46.2&\textbf{62.2}&60.7&71.1&25.4&80.4&4.8&26.9&81.8&0.0&0.0&-&42.0&2.2&39.1\\
  &&Ours&\textbf{88.1}&55.8&43.0&\textbf{29.3}&33.2&\textbf{50.4}&61.6&60.5&\textbf{75.3}&\textbf{43.8}&75.8&\textbf{6.5}&28.7&80.3&0.0&5.1&-&\textbf{46.8}&\textbf{20.2}&\textbf{42.3}\\
  \cline{2-23}
  &\multirow{5}{*}{\rotatebox{90}{BiseNet}}&Cityscapes\cite{yu2018bisenet}&67.1&32.3&25.3&9.6&19.4&6.7&7.7&16.1&49.6&19.8&43.1&0.0&0.0&9.1&0.0&0.0&-&0.0&0.0&16.1\\
  &&Foggy Cityscapes\cite{sakaridis2018semantic}&72.8&35.1&38.6&11.1&\textbf{23.2}&13.0&34.4&28.9&\textbf{59.8}&\textbf{33.0}&66.4&0.0&15.9&21.5&0.0&0.0&-&21.3&0.0&25.0\\
  &&CycleGAN\cite{zhu2017unpaired}&\textbf{75.5}&\textbf{41.2}&20.1&\textbf{32.4}&12.9&25.6&54.7&49.6&58.6&25.2&23.0&0.0&\textbf{19.2}&57.6&0.0&0.0&-&15.0&\textbf{4.6}&27.1\\
  &&MUNIT\cite{huang2018munit}&58.6&22.6&\textbf{43.3}&2.3&6.3&18.8&46.5&41.3&48.7&13.1&\textbf{85.9}&0.1&15.8&\textbf{63.3}&0.0&\textbf{0.2}&-&27.9&0.0&26.0\\
  &&Ours&72.1&26.2&38.5&12.6&8.4&\textbf{26.2}&\textbf{56.6}&\textbf{52.0}&54.7&21.0&77.0&\textbf{0.2}&3.2&46.6&0.0&0.0&-&\textbf{45.0}&0.0&\textbf{28.4}\\
  \hline
  \multirow{10}{*}{\rotatebox{90}{Foggy Driving}}&\multirow{5}{*}{\rotatebox{90}{RefineNet}}&Cityscapes\cite{lin2017refinenet}&90.1&29.3&68.3&27.3&16.7&41.3&54.2&59.6&68.0&6.8&88.7&60.9&45.4&66.4&5.5&9.6&45.4&9.8&48.4&44.3\\
  &&Foggy Cityscapes\cite{sakaridis2018semantic}&91.7&29.7&73.0&29.0&14.8&43.4&54.0&\textbf{61.6}&71.2&6.9&85.7&59.3&46.7&67.3&8.4&17.2&53.7&13.1&48.9&46.1\\
  &&CycleGAN\cite{zhu2017unpaired}&\textbf{92.4}&\textbf{33.2}&\textbf{73.3}&16.7&18.2&\textbf{45.8}&\textbf{55.2}&56.1&75.7&8.8&90.8&\textbf{65.1}&49.3&67.9&8.5&22.5&\textbf{78.0}&0.0&\textbf{49.2}&47.7\\
  &&MUNIT\cite{huang2018munit}&92.1&31.1&72.5&26.8&10.7&43.9&52.6&53.4&70.7&7.6&\textbf{92.1}&60.8&\textbf{49.7}&70.1&7.3&\textbf{35.7}&70.9&\textbf{16.9}&42.3&\textbf{47.8}\\
  &&Ours &91.9&28.2&71.7&\textbf{30.5}&\textbf{20.6}&44.4&48.5&57.8&\textbf{74.4}&\textbf{12.5}&90.2&61.7&49.6&\textbf{73.9}&\textbf{32.4}&15.9&52.8&0.5&45.8&\textcolor{red}{47.5}\\
  \cline{2-23}
  &\multirow{5}{*}{\rotatebox{90}{BiseNet}}&Cityscapes\cite{yu2018bisenet}&85.1&21.5&46.9&6.2&\textbf{13.1}&12.1&24.4&31.9&61.0&1.8&66.2&43.6&17.1&39.3&0.3&12.7&1.4&0.0&32.2&27.2\\
  &&Foggy Cityscapes\cite{sakaridis2018semantic}&\textbf{88.0}&23.7&\textbf{56.0}&\textbf{23.8}&7.4&16.2&31.9&32.7&\textbf{68.3}&0.8&79.1&42.2&16.4&50.8&0.2&13.7&5.8&0.0&18.4&30.3\\
  &&CycleGAN\cite{zhu2017unpaired}&81.9&\textbf{31.0}&28.2&6.7&11.3&29.6&41.0&\textbf{49.3}&47.5&2.2&19.2&\textbf{49.3}&\textbf{41.0}&57.3&\textbf{11.1}&10.1&15.8&0.0&\textbf{37.0}&30.0\\
  &&MUNIT\cite{huang2018munit}&73.9&10.4&44.4&8.9&3.1&28.0&41.1&35.0&58.2&1.8&78.4&39.2&38.7&\textbf{61.8}&3.0&4.7&17.0&3.7&28.1&30.5\\
  &&Ours &81.3&19.2&51.9&6.7&10.9&\textbf{33.5}&\textbf{46.5}&42.4&51.0&\textbf{2.9}&\textbf{85.0}&8.6&30.0&57.2&3.3&\textbf{13.8}&\textbf{19.1}&\textbf{5.7}&16.9&\textbf{30.8}\\
  \hline
 \end{tabular}
}
\caption{\label{sup:table_cityscapes} Results of semantic segmentation on the Foggy Zurich and Foggy Driving dataset based on RefineNet model with ResNet-101 backbone and BiseNet with ResNet-18 backbone using different simulated foggy images. The results are reported on mIoU over 19 categories. The best result is denoted in bold.}
\end{table*}

\begin{table*}[h]
\setlength{\tabcolsep}{3pt}
\centering
 \resizebox{\textwidth}{34mm}
 {
 \begin{tabular}{c|c|c|ccccccccccccccccccc|c}
 \hline
 \multicolumn{23}{c}{Virtual KITTI$\rightarrow$Synscapes}\\
  \hline
  \rotatebox{90}{Testing}&\rotatebox{90}{Model}&\rotatebox{90}{Fine-Tuning}&\rotatebox{90}{road}&\rotatebox{90}{sidewalk}&\rotatebox{90}{building}&\rotatebox{90}{wall}&\rotatebox{90}{fence}&\rotatebox{90}{pole}&\rotatebox{90}{traffic light}&\rotatebox{90}{traffic sign}&\rotatebox{90}{vegetation}&\rotatebox{90}{terrian}&\rotatebox{90}{sky}&\rotatebox{90}{person}&\rotatebox{90}{rider}&\rotatebox{90}{car}&\rotatebox{90}{truck}&\rotatebox{90}{bus}&\rotatebox{90}{train}&\rotatebox{90}{motorbike}&\rotatebox{90}{bicycle}&mIoU\\
  \hline
  \multirow{10}{*}{\rotatebox{90}{Foggy Zurich}}&\multirow{5}{*}{\rotatebox{90}{RefineNet}}&Cityscapes\cite{lin2017refinenet}&74.3&56.5&35.5&20.2&23.8&39.6&54.4&58.3&58.3&28.9&66.8&1.6&27.4&81.7&0.0&0.0&-&21.1&\textbf{6.2}&34.6\\
  &&Foggy Synscapes\cite{hahner2019semantic}&83.6&\textbf{60.0}&46.6&31.9&33.6&45.1&62.2&61.5&68.3&35.2&79.0&\textbf{4.3}&21.5&82.0&0.0&0.2&-&\textbf{44.7}&5.1&40.3\\
  &&CycleGAN\cite{zhu2017unpaired}&83.8&50.5&\textbf{69.1}&31.7&39.0&48.1&\textbf{62.4}&62.5&71.8&38.6&\textbf{92.4}&2.3&\textbf{29.2}&\textbf{83.0}&0.0&0.1&-&23.4&2.8&41.6\\
  &&MUNIT\cite{huang2018munit}&\textbf{85.4}&55.2&59.1&\textbf{40.4}&37.8&\textbf{49.2}&59.6&\textbf{62.8}&\textbf{72.0}&32.0&86.2&1.5&22.2&78.2&0.0&0.0&-&26.8&1.7&40.5\\
  &&Ours &83.7&55.7&56.3&40.0&\textbf{40.3}&45.1&61.5&59.1&70.5&\textbf{40.1}&88.1&4.1&24.9&79.8&0.0&\textbf{8.4}&-&33.9&2.3&\textbf{41.8}\\
  \cline{2-23}
  &\multirow{5}{*}{\rotatebox{90}{BiseNet}}&Cityscapes\cite{yu2018bisenet}&67.1&32.3&25.3&9.6&19.4&6.7&7.7&16.1&49.6&19.8&43.1&0.0&0.0&9.1&0.0&0.0&-&0.0&0.0&16.1\\
  &&Foggy Synscapes\cite{hahner2019semantic}&71.6&\textbf{36.6}&\textbf{52.4}&\textbf{28.8}&\textbf{25.6}&17.4&26.2&38.0&65.6&\textbf{38.8}&\textbf{87.7}&0.7&1.7&37.0&0.0&0.0&-&0.0&0.0&27.8\\
  &&CycleGAN\cite{zhu2017unpaired}&52.8&31.4&46.0&19.4&13.5&\textbf{32.5}&\textbf{56.7}&51.1&61.7&6.2&83.9&2.2&\textbf{25.2}&34.9&0.0&0.0&-&\textbf{55.5}&\textbf{13.9}&30.9\\
  &&MUNIT\cite{huang2018munit}&\textbf{72.9}&31.6&29.5&17.8&13.8&26.9&54.8&46.5&\textbf{67.9}&1.1&59.9&0.5&22.7&\textbf{68.7}&0.0&0.0&-&8.4&0.0&27.5\\
  &&Ours&44.7&27.9&43.5&23.7&15.1&27.8&47.7&\textbf{55.6}&61.5&32.4&81.5&\textbf{2.9}&10.2&63.1&0.0&0.0&-&51.6&9.1&\textbf{31.5}\\
  \hline
  \multirow{10}{*}{\rotatebox{90}{Foggy Driving}}&\multirow{5}{*}{\rotatebox{90}{RefineNet}}&Cityscapes\cite{lin2017refinenet}&90.1&29.3&68.3&\textbf{27.3}&16.7&41.3&54.2&59.6&68.0&6.8&88.7&\textbf{60.9}&45.4&66.4&5.5&9.6&45.4&9.8&48.4&44.3\\
  &&Foggy Synscapes\cite{hahner2019semantic}&\textbf{92.4}&32.9&76.1&16.8&14.6&43.3&\textbf{55.0}&\textbf{60.8}&74.0&9.3&90.8&49.8&36.0&72.2&17.5&\textbf{51.3}&65.0&11.1&\textbf{50.3}&48.4\\
  &&CycleGAN\cite{zhu2017unpaired}&91.0&30.0&73.7&11.9&17.2&45.8&52.2&56.4&72.5&9.0&88.9&56.8&40.0&\textbf{75.6}&17.6&39.2&62.0&\textbf{26.9}&42.5&47.8\\
  &&MUNIT\cite{huang2018munit}&91.6&\textbf{35.7}&\textbf{76.5}&19.9&\textbf{18.5}&\textbf{46.4}&53.2&59.5&75.1&8.9&\textbf{92.5}&48.6&40.4&73.3&18.7&40.8&65.6&14.3&39.0&48.3\\
  &&Ours &92.1&35.0&74.1&20.7&17.5&40.0&51.1&57.3&\textbf{75.4}&\textbf{10.8}&91.3&53.8&\textbf{48.3}&74.7&\textbf{21.3}&45.2&\textbf{75.5}&12.4&49.1&\textbf{49.8}\\
  \cline{2-23}
  &\multirow{5}{*}{\rotatebox{90}{BiseNet}}&Cityscapes\cite{yu2018bisenet}&\textbf{85.1}&21.5&46.9&6.2&\textbf{13.1}&12.1&24.4&31.9&61.0&1.8&66.2&43.6&17.1&39.3&0.3&12.7&1.4&0.0&32.2&27.2\\
  &&Foggy Synscapes\cite{hahner2019semantic}&81.4&16.6&\textbf{60.8}&5.3&8.7&27.4&33.8&43.7&60.1&\textbf{2.8}&\textbf{91.9}&31.0&5.5&57.2&11.4&22.5&8.0&0.0&18.6&30.9\\
  &&CycleGAN\cite{zhu2017unpaired}&62.1&15.2&57.1&5.4&2.4&\textbf{33.3}&\textbf{42.7}&\textbf{47.8}&66.3&0.7&85.7&\textbf{45.1}&\textbf{34.0}&45.8&9.0&25.8&7.9&\textbf{21.4}&21.0&33.1\\
  &&MUNIT\cite{huang2018munit}&80.5&\textbf{24.2}&57.4&\textbf{15.6}&6.1&29.3&39.6&46.6&\textbf{70.5}&1.8&85.8&29.7&7.9&57.7&\textbf{11.7}&24.2&8.2&0.1&27.1&32.8\\
  &&Ours &61.6&17.3&58.7&6.1&2.5&25.6&37.9&41.8&68.4&1.8&91.4&42.8&23.3&\textbf{64.3}&4.9&\textbf{31.8}&\textbf{25.7}&4.5&\textbf{39.3}&\textbf{34.2}\\
  \hline
 \end{tabular}
}
\caption{\label{sup:table_synscapes} Results of semantic segmentation on the Foggy Zurich and Foggy Driving dataset based on RefineNet model with ResNet-101 backbone and BiseNet with ResNet-18 backbone using different simulated foggy images. The results are reported on mIoU over 19 categories. The best result is denoted in bold.}
\end{table*}

\end{document}